\documentclass[runningheads]{llncs}

\usepackage[mobile]{eccv}

\usepackage{eccvabbrv}

\usepackage{graphicx}
\usepackage{booktabs}

\usepackage[accsupp]{axessibility}  

\usepackage{kotex}
\usepackage{wrapfig} 
\usepackage{tcolorbox}
\usepackage{multirow}
\usepackage{longtable}
\usepackage{algorithm}
\usepackage{algorithmic}

\usepackage[hypertexnames=false]{hyperref}

\usepackage{orcidlink}

\begin{document}

\title{Orthogonal Negative Guidance in Attention Feature Space for Text-to-Image Generation}

\titlerunning{Orthogonal Negative Guidance in Attention Feature Space}

\author{Jungmin Ko\inst{1} \and
Jungwon Park\inst{2,5} \and
Jimyeong Kim\inst{3} \and 
Changin Choi\inst{1,6} \and
Wonseok Lee\inst{1} \and
Wonjong Rhee\inst{1,4}
}

\authorrunning{J.~Ko et al.}

\institute{Interdisciplinary Program in Artificial Intelligence, Seoul National University 
\and
Research Institute for Convergence Science, Seoul National University
\and
Artificial Intelligence Institute, Seoul National University
\and
Department of Intelligence and Information, Seoul National University
\and
Daegu Gyeongbuk Institute of Science and Technology
\and
Samsung Advanced Institute of Technology, Samsung Electronics Co., Ltd\\
\email{\{jungminko, quoded97, wlaud1001, ci2015.choi, dnjstjr1017, wrhee\}@snu.ac.kr}}

\maketitle
\begin{figure}[]
\vspace{-7mm}
\begin{center}
   \includegraphics[width=0.85\linewidth]{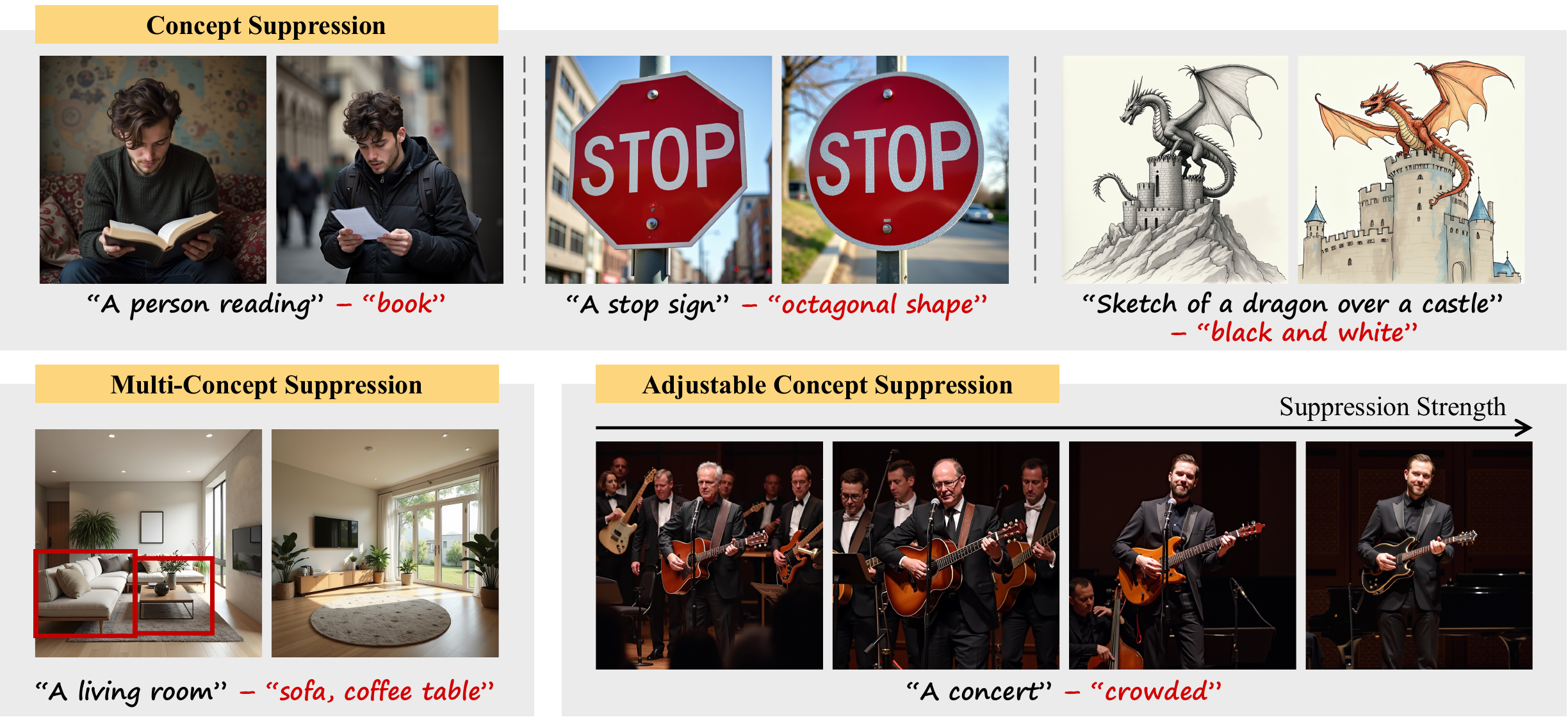}   
\end{center}
\vspace{-5mm}
\caption{
    Images generated without negative guidance~(left image of each pair) and with our Orthogonal Negative Guidance~(right image of each pair) on FLUX-dev. Black and red text indicate the positive and negative prompts, respectively. Our method effectively suppresses unwanted concepts across diverse scenarios and supports multi-concept suppression~(bottom left) and adjustable concept suppression~(bottom right).
}
\vspace{-10mm}
\label{fig:intro}
\end{figure}

\begin{abstract}
Text-to-image~(T2I) models have become increasingly capable of generating high-quality images. Yet, enforcing the explicit \textit{absence} of a specified object or attribute remains a fundamentally challenging problem. Existing approaches, including prompt negation, post-hoc editing, and negative guidance, remain insufficient for explicit concept suppression, often failing to remove the target concept or degrading overall image quality. To this end, we propose Orthogonal Negative Guidance in attention feature space, a training-free method that operates in the attention output space of MM-DiT-based T2I transformers. Our method orthogonalizes negative-prompt attention features with respect to positive-prompt features and subtracts only the orthogonal component, suppressing unwanted concepts while preserving desired semantics. Experiments on FLUX-dev and FLUX-schnell show that our method achieves favorable trade-offs between concept suppression, prompt alignment, and image quality. In human evaluation, our method outperforms the second-best baseline by 18.78\%. We further show that our method supports multi-concept suppression and adjustable concept suppression.
\end{abstract}

   

\section{Introduction}
\label{sec:introduction}
Text-to-image~(T2I) generative models have become increasingly capable of synthesizing high-quality images that faithfully reflect user-provided prompts. Despite this progress, controlling the absence of a concept, i.e., an object or an attribute, at inference time remains a persistent challenge. In many applications, such as safety filtering~\cite{schramowski2023safe,rando2022red, yang2024sneakyprompt}, debiasing~\cite{parihar2024balancing,li2025fair}, or fine-grained content control~\cite{gandikota2023erasing, Kumari_2023_ICCV, gandikota2024unified}, users may wish to explicitly exclude certain objects or attributes from the generated image without modifying model parameters. Enabling reliable, training-free suppression of unwanted concepts is therefore a critical yet unresolved challenge in controllable T2I generation.

In practice, several strategies can be used to exclude unwanted concepts. Prompt-level negation (e.g., inserting `without' or `no' into prompts) is the simplest way to instruct models to avoid certain concepts; however, such approaches have been reported to generate the negated concept even more frequently than prompts without explicit negation~\cite{li2024get,alhamoud2025vision}. Another strategy is to adopt generate-and-edit pipelines, which first synthesize an image from the prompt and then apply an image editing model to remove undesired concepts. These editing models~\cite{labs2025flux, wu2025qwenimagetechnicalreport, wu2025omnigen2} are typically trained to preserve spatial and structural information from the input image. As a result, they are inherently constrained in performing large structural modifications. This training objective conflicts with the removal of concepts that play a central role in the scene, which often requires broader restructuring rather than simple local substitutions. Consequently, while generate-and-edit pipelines show promising performance on certain editing tasks, they are less effective for substantial concept removal. In such cases, they either fail to eliminate the concept or produce results that are semantically inconsistent with the original prompt. These limitations highlight the need for mechanisms that more explicitly enforce concept suppression during image generation.

Negative guidance~\cite{ho2022classifier, chen2025normalized, nguyen2025supercharged, guo2025vsf} aims to provide such generation-time control by combining positive and negative conditioning signals. Techniques such as classifier-free guidance (CFG)~\cite{ho2022classifier} and related variants~(e.g., NAG~\cite{chen2025normalized}) steer the generation process away from undesired concepts through negative prompts. However, these methods typically apply negative guidance signals through linear subtraction without explicitly separating components aligned with the desired semantics. As a result, suppression may degrade prompt alignment before sufficiently eliminating the unwanted concept. To address this limitation, we propose \textit{Orthogonal Negative Guidance} in attention feature space for rectified flow T2I models. Our method intervenes in the image-to-text attention outputs within MM-DiT blocks, which mediate cross-modal interactions between text and image tokens. For each block, we compute two attention outputs: one conditioned on the positive prompt~(i.e., the original generation prompt) and one conditioned on an auxiliary negative prompt specifying the concept to suppress. We treat the image-to-text attention output from the negative-conditioned branch as a guidance signal. Instead of directly subtracting it from the positive attention output, we remove only the component of this signal that is orthogonal to the corresponding positive attention output. In this way, the method effectively suppresses information specific to the unwanted concept while preserving semantic directions aligned with the positive prompt. The method is training-free, introduces no additional parameters, and can be seamlessly integrated into recently proposed rectified-flow transformer architectures.

Beyond methodological limitations, existing negative guidance methods are primarily evaluated on improving perceptual quality using negative prompts such as `blurry' or `low quality', rather than on diverse concept suppression scenarios. Consequently, their effectiveness in removing specific concepts while preserving prompt alignment remains insufficiently examined. To address this evaluation gap, we introduce the \textit{Diverse Concept Suppression Benchmark}~(DCS-Bench), covering six categories of concept suppression scenarios in T2I generation. 

We evaluate our method on rectified-flow T2I models, including FLUX-dev and FLUX-schnell~\cite{flux2024}, using the proposed benchmark. Quantitatively, our approach achieves favorable trade-offs between prompt alignment and concept suppression, demonstrating effective suppression without sacrificing alignment with the positive prompt. It also achieves favorable trade-offs between image quality and concept suppression, indicating minimal quality degradation. In human evaluation, our method achieves the best result, outperforming the second-best approach by 18.78\% in preference scores. These results demonstrate that Orthogonal Negative Guidance substantially improves concept suppression while preserving both prompt alignment and visual quality. Finally, we show that our method supports continuous fader control via guidance scaling and can suppress multiple unwanted concepts simultaneously.

\section{Related Work}
\label{sec:related_work}

\subsection{Controllable Text-to-Image Generation}
With the rapid advancement of text-to-image foundation models such as Stable Diffusion~\cite{rombach2022high, podell2023sdxl, sd3.5} and FLUX~\cite{flux2024}, numerous methods have been developed to enhance generation controllability. These approaches address a wide range of tasks, including image editing~\cite{hertz2022prompt, brooks2023instructpix2pix, tumanyan2023plug, meng2021sdedit, labs2025flux, wu2025qwenimagetechnicalreport, kim2025reflex, kulikov2025flowedit}, personalization~\cite{ruiz2023dreambooth, gal2022image, kumari2023multi,han2023svdiff, avrahami2023break, kim2024selectively}, spatially conditioned generation~\cite{zhang2023adding, li2023gligen, bar2023multidiffusion, xie2023boxdiff, mou2024t2i}, and fine-grained semantic control~\cite{chefer2023attend, park2024cross, brack2023sega, gandikota2024concept, liu2022compositional, byun2025directional}. In particular, fine-grained control methods aim to enforce or suppress specific concepts during inference. Among them, negative guidance provides an inference-time mechanism for steering generation away from undesired concepts by incorporating negative conditioning signals into the denoising process. Due to its simplicity and compatibility with existing T2I pipelines, negative guidance is widely adopted in practice to suppress artifacts or unwanted attributes, such as `low quality,' `missing fingers,' or `jpeg artifacts.'

\subsection{Negative Guidance for Text-to-Image Generative Models}
Negative guidance was first introduced by modifying classifier-free guidance (CFG)~\cite{ho2022classifier}, replacing the unconditional prediction with a prediction conditioned on an auxiliary negative prompt. Although simple and training-free, its effectiveness can degrade when the positive- and negative-conditioned predictions diverge substantially, particularly in few-step sampling settings such as FLUX-schnell. To mitigate this issue, subsequent works apply negative guidance in attention feature space rather than on the model output. For example, NASA~\cite{nguyen2025supercharged} applies negative guidance to cross-attention outputs in U-Net-based T2I models. NAG~\cite{chen2025normalized} introduces normalization and refinement steps to stabilize guidance, and VSF~\cite{guo2025vsf} modifies attention value vectors for efficient inference-time suppression. Despite these improvements, existing approaches typically apply negative signals through linear subtraction without explicitly separating components aligned with the desired semantics, which may degrade alignment with the positive prompt before sufficiently suppressing the unwanted concept. In contrast, our method isolates and subtracts only the components orthogonal to the semantic directions induced by the positive prompt, enabling effective suppression while preserving prompt alignment.

\section{Method}
\label{sec:method}
\begin{figure}[t]
\centering
\begin{subfigure}[t]{0.68\columnwidth}
    \centering
    \includegraphics[width=\linewidth]{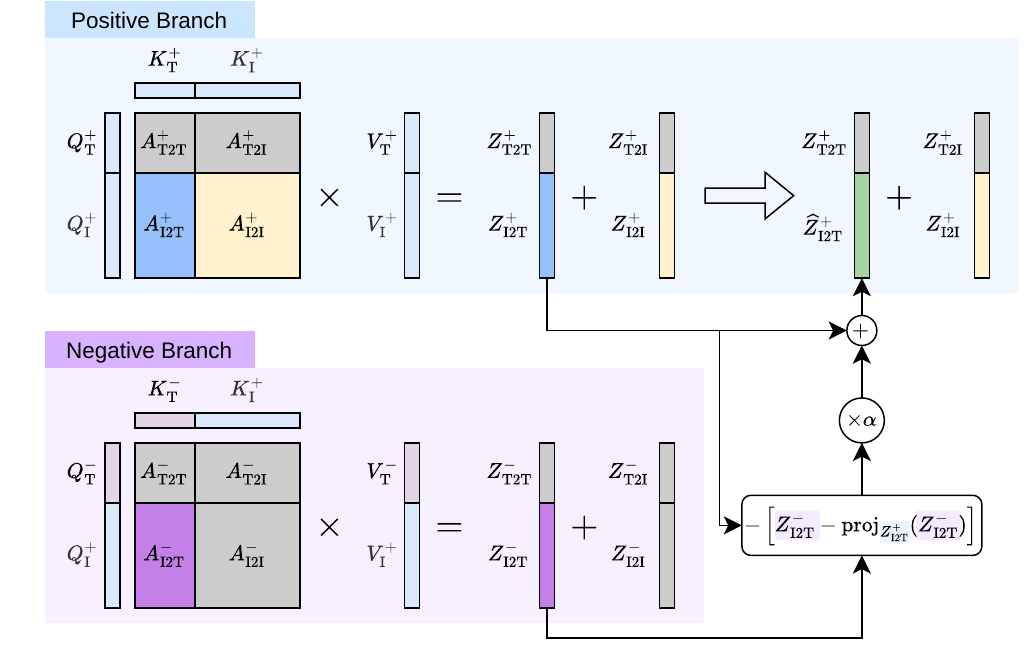}
    \caption{}
    \label{fig:method1}
\end{subfigure}
\begin{subfigure}[t]{0.30\columnwidth}
    \centering
    \includegraphics[width=\linewidth]{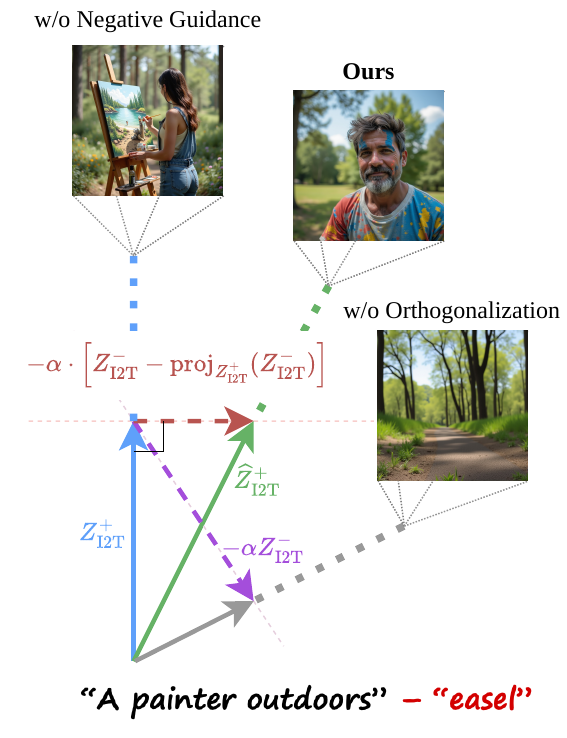}
    \caption{}
    \label{fig:method2}
\end{subfigure}
\caption{\textbf{Orthogonal Negative Guidance in attention feature space.} (a) Overview of Orthogonal Negative Guidance in each MM-DiT attention block. 
(b) Illustration of three cases:  without negative guidance, Orthogonal Negative Guidance (ours), and negative guidance without orthogonalization.
The generated images for an exemplary case are shown.
Orthogonal Negative Guidance effectively suppresses the generation of unwanted concept (easel) while preserving the prompt-consistent scene.
}
\vspace{-3mm}
\label{fig:method}
\end{figure}

Our method introduces \emph{orthogonal negative guidance} in attention feature space. Attention features induced by a negative prompt are orthogonalized with respect to those induced by a positive prompt and then subtracted, effectively suppressing the generation of unwanted concepts while minimizing interference to the desired concept. 
An overview of our method is shown in Figure~\ref{fig:method}.

\subsection{MM-DiT Attention in Rectified Flow Models}
Rectified Flow~\cite{liu2022flow, albergo2022building, lipman2022flow} transformer models such as FLUX~\cite{blackforestlabs} and Stable Diffusion 3.5 Large~\cite{sd3.5} employ multimodal diffusion transformer~(MM-DiT)~\cite{peebles2023scalable, esser2024scaling} architectures, in which each MM-DiT block jointly processes text and image features via multi-head attention~\cite{vaswani2017attention}. 
Let $X_\mathrm{T}$ and $X_\mathrm{I}$ denote the input text and image features to an MM-DiT block, respectively. Within each attention head, features from each modality are linearly projected into query, key, and value representations: $ Q_{\mathrm{T}}, K_{\mathrm{T}}, V_{\mathrm{T}} $ for the text modality and $ Q_{\mathrm{I}}, K_{\mathrm{I}}, V_{\mathrm{I}} $ for the image modality. 
The projections from both modalities are concatenated along the token dimension to form 
\begin{equation}
    Q=
    \begin{bmatrix} Q_{\mathrm{T}} \\
    Q_{\mathrm{I}}
    \end{bmatrix} \in \mathbb{R}^{N \times d_k}, \quad
    K=
    \begin{bmatrix} K_{\mathrm{T}} \\
    K_{\mathrm{I}}
    \end{bmatrix} \in \mathbb{R}^{N \times d_k}, \quad
    V=
    \begin{bmatrix} V_{\mathrm{T}} \\
    V_{\mathrm{I}}
    \end{bmatrix} \in \mathbb{R}^{N \times d_v},
\end{equation}
where $N$ is the token dimension~(i.e., the number of tokens) and $d_k$, $d_v$ are the key and hidden dimensions, respectively.
The attention output is then computed as
\begin{equation}
    Z=AV \in \mathbb{R}^{N \times d_v}, \quad
    A=\mathrm{softmax}\!\left(\frac{QK^\top}{\sqrt{d_k}}\right) \in \mathbb{R}^{N \times N}.
\end{equation}
To make cross-modal interactions explicit, this attention computation can be rewritten by partitioning it according to modality. Specifically, the attention output $Z$ can be expressed as the product of a block-structured attention matrix and the corresponding value vectors:
\begin{equation} 
    Z = 
    \begin{bmatrix} Z_{\mathrm{T}} \\ 
    Z_{\mathrm{I}} 
    \end{bmatrix} = 
    \begin{bmatrix} 
    A_{\mathrm{T2T}} & A_{\mathrm{T2I}} \\ 
    A_{\mathrm{I2T}} & A_{\mathrm{I2I}} 
    \end{bmatrix} 
    \begin{bmatrix} V_{\mathrm{T}}\\
    V_{\mathrm{I}}
    \end{bmatrix},
\label{eq:MM-DiT attention}
\end{equation}
Here, the full attention map $A$ is obtained by applying the softmax operation to the block-wise similarity scores between queries and keys from the two modalities:
\begin{equation}
    A = 
    \begin{bmatrix} 
    A_{\mathrm{T2T}} & A_{\mathrm{T2I}} \\ 
    A_{\mathrm{I2T}} & A_{\mathrm{I2I}} 
    \end{bmatrix} =
    \mathrm{softmax}\!\left(
    \frac{1}{\sqrt{d_k}}
    \begin{bmatrix}
    Q_{\mathrm{T}} K_{\mathrm{T}}^\top &
    Q_{\mathrm{T}} K_{\mathrm{I}}^\top \\
    Q_{\mathrm{I}} K_{\mathrm{T}}^\top &
    Q_{\mathrm{I}} K_{\mathrm{I}}^\top
    \end{bmatrix}
    \right).
\end{equation}
The submatrices $ A_{\mathrm{T2T}}$, $A_{\mathrm{T2I}}$, $A_{\mathrm{I2T}}$, and $A_{\mathrm{I2I}}$ represent attentions from query tokens of one modality to key tokens of another. Among these components, image-to-text attention~($A_{\mathrm{I2T}}$) is known to capture text-image relationships that encode semantic alignment between modalities, while image-to-image attention~($A_{\mathrm{I2I}}$) captures spatial layout and structural information within the image~\cite{kim2025reflex, shin2025exploring}. These roles are analogous to those of cross-attention and self-attention in U-Net-based text-to-image models~\cite{rombach2022high, hertz2022prompt, tumanyan2023plug}. Motivated by this role of $A_{\mathrm{I2T}}$, our method operates on the corresponding image-to-text attention output, which is the value-weighted aggregation of $A_{\mathrm{I2T}}$, within MM-DiT blocks for negative guidance.


\subsection{Orthogonal Negative Guidance in Attention Feature Space}
Let $P^{+}$ and $P^{-}$ denote the positive and negative prompts, respectively. We use superscripts $+$ and $-$ to indicate quantities computed under each prompt.
For each attention head, the attention output under the positive prompt is computed as 
\begin{equation} 
    Z^{+}= 
    \begin{bmatrix} Z_{\mathrm{T}}^{+} \\ 
    Z_{\mathrm{I}}^{+} 
    \end{bmatrix} = 
    \begin{bmatrix} 
    A_{\mathrm{T2T}}^{+} & A_{\mathrm{T2I}}^{+} \\ 
    A_{\mathrm{I2T}}^{+} & A_{\mathrm{I2I}}^{+} 
    \end{bmatrix} 
    \begin{bmatrix} V_{\mathrm{T}}^{+} \\
    V_{\mathrm{I}}^{+}
    \end{bmatrix}.
\end{equation}
This block formulation induces the modality-wise decomposition
\begin{equation} 
    \begin{aligned}
        Z_{\mathrm{T}}^{+} &= A_{\mathrm{T2T}}^{+} V_{\mathrm{T}}^{+} + A_{\mathrm{T2I}}^{+} V_{\mathrm{I}}^{+} = Z_{\mathrm{T2T}}^{+} + Z_{\mathrm{T2I}}^{+}, \\
        Z_{\mathrm{I}}^{+} &= A_{\mathrm{I2T}}^{+} V_{\mathrm{T}}^{+} + A_{\mathrm{I2I}}^{+} V_{\mathrm{I}}^{+} = Z_{\mathrm{I2T}}^{+} + Z_{\mathrm{I2I}}^{+}.
    \end{aligned}
\end{equation}
To ensure spatial alignment between branches, the negative branch reuses the image-side features $(Q_{\mathrm{I}}^{+}, K_{\mathrm{I}}^{+}, V_{\mathrm{I}}^{+})$ from the positive branch while retaining its own text-side features $(Q_{\mathrm{T}}^{-}, K_{\mathrm{T}}^{-}, V_{\mathrm{T}}^{-})$. The attention output under the negative prompt is then computed as
\begin{equation}
\label{eqn:negative branch}
    Z^{-}= 
    \begin{bmatrix} Z_{\mathrm{T}}^{-} \\ 
    Z_{\mathrm{I}}^{-} 
    \end{bmatrix} = 
    \begin{bmatrix} 
    A_{\mathrm{T2T}}^{-} & A_{\mathrm{T2I}}^{-} \\ 
    A_{\mathrm{I2T}}^{-} & A_{\mathrm{I2I}}^{-} 
    \end{bmatrix} 
    \begin{bmatrix} V_{\mathrm{T}}^{-} \\
    V_{\mathrm{I}}^{+}
    \end{bmatrix},
\end{equation}
where $A^{-}$ is computed from $(Q_{\mathrm{T}}^{-}, K_{\mathrm{T}}^{-})$ together with $(Q_{\mathrm{I}}^{+}, K_{\mathrm{I}}^{+})$.
The image-modality output $Z_{\mathrm{I}}^{-}$ decomposes as
\begin{equation}
    Z_{\mathrm{I}}^{-}  = Z_{\mathrm{I2T}}^{-} + Z_{\mathrm{I2I}}^{-}, \quad
    Z_{\mathrm{I2T}}^{-} = A_{\mathrm{I2T}}^{-} V_{\mathrm{T}}^{-}.
\end{equation}
Sharing the image-side features between branches anchors the negative guidance signal to the same spatial representation as the positive branch. Without this constraint, the negative branch would operate on an independent image latent, leading to spatially inconsistent guidance, as shown in Figure~\ref{fig:ablation_attention_latent_sharing}. 
Our method applies negative guidance to the image-to-text attention output $Z_{\mathrm{I2T}}^{+}$ in the positive branch. To suppress the generation of unwanted concepts specified by $P^{-}$ while preserving semantic information aligned with $P^{+}$, we subtract only the component of $Z_{\mathrm{I2T}}^{-}$ that is orthogonal to $Z_{\mathrm{I2T}}^{+}$: 
\begin{equation}
\label{eqn:Orthogonal Guidance}
    \widehat{Z}_{\mathrm{I2T}}^{+} = Z_{\mathrm{I2T}}^{+} - \alpha \cdot \left[ Z_{\mathrm{I2T}}^{-} - \mathrm{proj}_{Z_{\mathrm{I2T}}^{+}} (Z_{\mathrm{I2T}}^{-}) \right]
\end{equation}
where $\alpha$ is the guidance scale and $\mathrm{proj}_{Z_{\mathrm{I2T}}^{+}} (\cdot)$ denotes projection onto $Z_{\mathrm{I2T}}^{+}$ along the hidden dimension. The projection is computed independently for each token and attention head. Therefore, the modified attention output for the positive branch becomes
\begin{equation}
    \widehat{Z}^{+}= 
    \begin{bmatrix} Z_{\mathrm{T}}^{+} \\ 
    \widehat{Z}_{\mathrm{I}}^{+} 
    \end{bmatrix} = 
    \begin{bmatrix} 
    Z_{\mathrm{T2T}}^{+} + Z_{\mathrm{T2I}}^{+} \\ 
    \widehat{Z}_{\mathrm{I2T}}^{+} + Z_{\mathrm{I2I}}^{+}
    \end{bmatrix}.
\end{equation}
The negative branch remains unchanged from Eq.(\ref{eqn:negative branch}). 
This procedure is applied to all MM-DiT blocks for timesteps $t \geq  \tau$, and the final image latent $\widehat{Z}_{\mathrm{I}}^{+}$ from the positive branch is decoded to generate the output image. The pseudocode for our method is provided in Algorithm~\ref{alg:orthogonal_negative_guidance} of Appendix~\ref{sec:appendix_method_pseudo_code}.

\begin{figure}[t]
\begin{center}
   \includegraphics[width=0.9\linewidth]{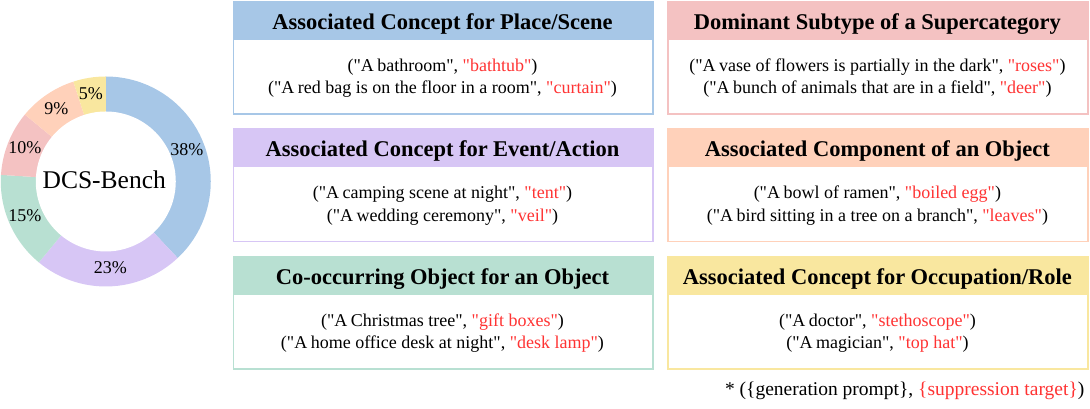}   
    \vspace{-6mm}
\end{center}
\caption{
    \textbf{Diverse Concept Suppression Benchmark~(DCS-Bench)}. The donut chart shows the proportion of each category in DCS-Bench. Each category captures a distinct semantic relationship between the generation prompt and the suppression target. Two representative suppression scenarios are shown per category. 
}
\vspace{-4mm}
\label{fig:benchmark}
\end{figure}

\vspace{-2mm}
\section{Diverse Concept Suppression Benchmark} 
\label{sec:benchmark}
Previous evaluations of negative guidance methods have primarily focused on improving perceptual quality, rather than systematically evaluating performance across diverse concept suppression scenarios. 
A recent work of VSF~\cite{guo2025vsf} considers beyond simple perceptual quality and evaluates cases where the prompt specifies an object and the negative prompt specifies one of its components to be suppressed. However, its evaluation is largely limited to this specific setting and does not cover a broader range of semantic relationships.
%
To address this limitation, we introduce the Diverse Concept Suppression Benchmark~(DCS-Bench), which evaluates negative guidance methods across various semantic relationships between a generation prompt and a target concept to be suppressed. To construct the benchmark, we identify prompts for which a state-of-the-art T2I model frequently generates additional concepts that are not explicitly specified in the prompt. These recurrent but unspecified concepts serve as natural suppression targets because they frequently appear in generated images despite not being requested in the prompt. We use large language models~(LLMs) and vision-language models~(VLMs) to collect such (\{prompt\}, \{suppression target\}) pairs, resulting in 200 suppression scenarios. Details of the dataset construction process are provided in Appendix~\ref{sec:appendix_benchmark_construction}.

We group these 200 suppression scenarios into six categories based on the semantic relationship between the prompt and the suppression target: (1) Associated Concept for Place/Scene, (2) Associated Concept for Event/Action, (3) Co-occurring Object for an Object, (4) Dominant Subtype of a Supercategory, (5) Associated Component of an Object, and (6) Associated Concept for Occupation/Role. Figure~\ref{fig:benchmark} presents representative examples of each category and their proportions in the benchmark. The first category, Associated Concept for Place/Scene, consists of scenarios where the prompt describes a place or scene and the suppression target is a concept commonly associated with that place/scene, which often causes T2I models to generate the concept even when it is not mentioned in the prompt. For instance, images generated from `A bathroom' often include a `bathtub' despite the absence of that term in the prompt. The remaining categories are defined similarly, with detailed descriptions deferred to Appendix~\ref{sec:appendix_benchmark_types}.

Using DCS-Bench, negative guidance methods can be evaluated by generating images from the prompt while providing the frequently generated concept as a negative prompt, with the goal of suppressing that concept during generation. Performance is measured by how effectively the target concept is suppressed while preserving prompt alignment and image quality. To our knowledge, DCS-Bench is the first benchmark designed to evaluate negative guidance methods across diverse concept suppression scenarios.

\section{Experiments}
\label{sec:experiments}

\subsection{Experimental Setup}

\subsubsection{Implementation Details.}
We evaluate our method on several rectified flow T2I models, including FLUX-dev~\cite{flux2024}, FLUX-schnell~\cite{flux2024}, and Stable Diffusion 3.5 Large~(SD3.5-Large)~\cite{sd3.5}. For FLUX-dev and SD3.5-Large, we use 28 sampling steps, guidance scale $\alpha=4$, and apply negative guidance starting from timestep $\tau=2$. For FLUX-schnell, a few-step generation model designed for faster inference with slightly lower image quality, we use 4 sampling steps, guidance scale $\alpha=2$, and apply negative guidance starting from timestep $\tau=0$. Unless otherwise specified, all results are generated using FLUX-dev. Additional implementation details are provided in Appendix~\ref{sec:appendix_implementation_details}.

\begin{figure*}[t]
\begin{center}
   \includegraphics[width=0.75\linewidth]{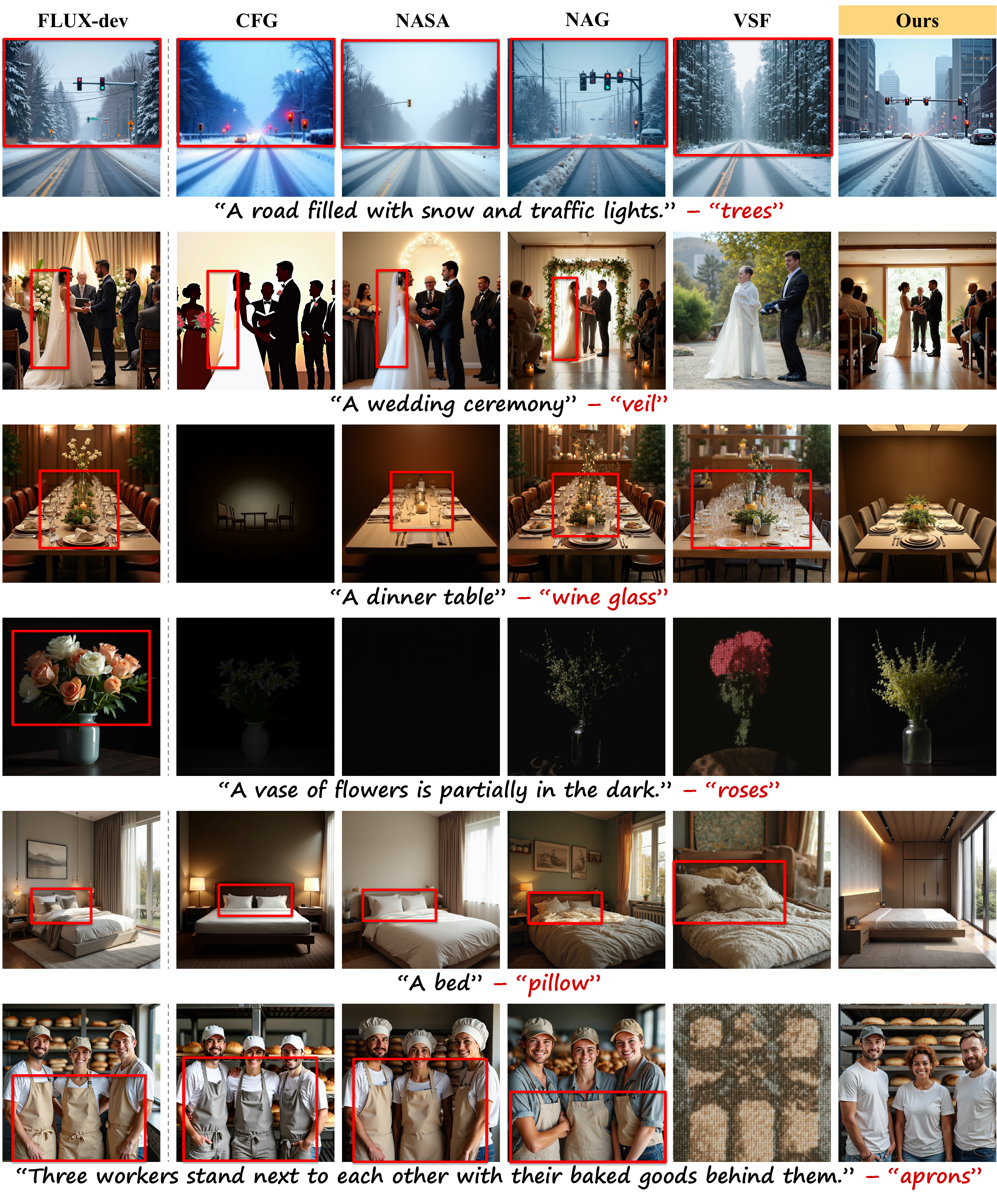}
    \vspace{-6mm}
\end{center}
\caption{
    \textbf{Qualitative comparison with negative guidance baselines on FLUX-dev.}
}
\vspace{-8mm}
\label{fig:qualitative_comparison}
\end{figure*}

\subsubsection{Evaluation Metrics.}
We adopt three VLM-based metrics to evaluate concept suppression performance: Negative Concept Suppression, Prompt Alignment, and Image Quality. Negative Concept Suppression measures whether the target concept is successfully suppressed by asking the VLM to determine whether the concept is absent from the generated image. Prompt Alignment and Image Quality are measured similarly by prompting the VLM to score how well the image aligns with the prompt and to assess its visual quality. We use Qwen3-VL-8B~\cite{Qwen3-VL} as the evaluator VLM. To assess the reliability of these metrics, we compare the VLM evaluations with human annotations. The results show 91\% agreement for Negative Concept Suppression and strong Pearson correlations~($r \geq 0.71$) for both Prompt Alignment and Image Quality, indicating that the VLM-based metrics align well with human judgements. The exact instructions used for each metric and additional details on the human annotation comparison are provided in Appendix~\ref{sec:appendix_evaluation}.

\subsubsection{Baseline Methods for Comparison.}
We compare our method with existing negative guidance methods and generate-and-edit pipelines that use image editing models. The negative guidance baselines include CFG~\cite{ho2022classifier}, where the unconditional prediction is replaced with a prediction conditioned on an auxiliary negative prompt, NASA~\cite{nguyen2025supercharged}, NAG~\cite{chen2025normalized}, and VSF~\cite{guo2025vsf}. All these methods use the suppression target as the negative prompt. For generate-and-edit baselines, we consider recently proposed image editing models including Kontext~\cite{labs2025flux}, Qwen-Image-Edit~\cite{wu2025qwenimagetechnicalreport}, and OmniGen2~\cite{wu2025omnigen2}. These pipelines first generate images using the same T2I model as ours without negative prompts, and then apply the editing models to remove the suppression target from the generated images. For these models, we use `Remove \{suppression target\}' as the editing instruction, while results obtained using alternative editing instructions are provided in Appendix~\ref{sec:appendix_editing_instruction}. Additional implementation details for all baselines are provided in Appendix~\ref{sec:appendix_implementation_details}.

\begin{figure*}[t]
\begin{center}
   \includegraphics[width=0.70\linewidth]{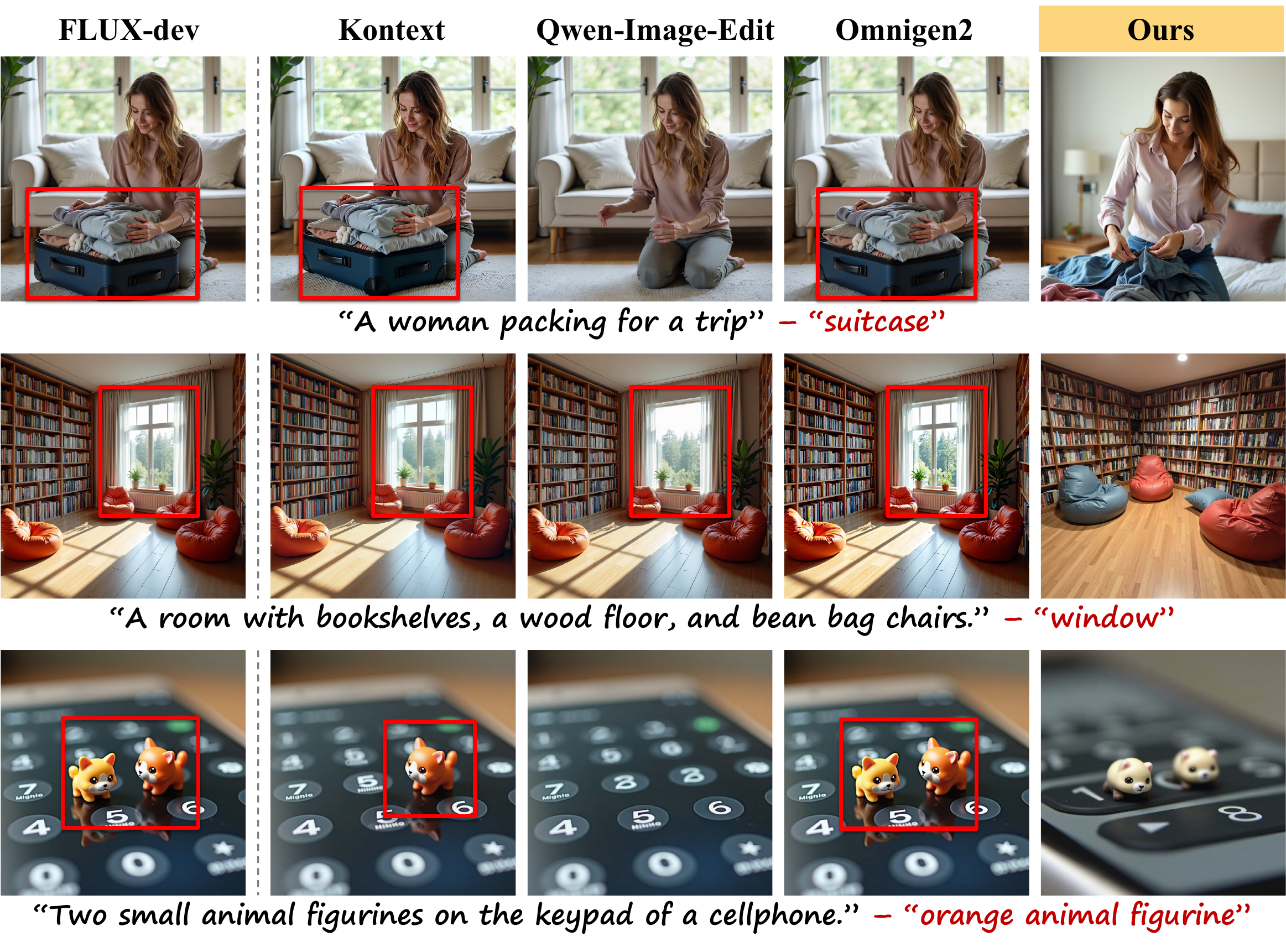} 
    \vspace{-5mm}
\end{center}
\caption{
    \textbf{Qualitative comparison with generate-and-edit baselines on FLUX-dev.}
}
\vspace{-5mm}
\label{fig:qualitative_editing}
\end{figure*}

\subsection{Experimental Results}
\subsubsection{Qualitative Comparison.}
Figure~\ref{fig:qualitative_comparison} compares our method with other negative guidance methods across all six categories in DCS-Bench. As shown in the figure, prior methods frequently fail to suppress the target concepts, and some methods such as CFG and VSF often produce images with degraded quality. In contrast, our method successfully suppresses the target concepts while preserving both prompt alignment and image quality. For example, in the first row of Figure~\ref{fig:qualitative_comparison}, all other negative guidance methods fail to suppress the generation of trees, and CFG and NASA produce blurry images indicating degraded image quality. Our method, however, successfully suppresses the generation of trees while maintaining both prompt alignment and image quality. Figure~\ref{fig:qualitative_editing} compares our method with generate-and-edit pipelines based on various image editing models. Because these editing models are trained to preserve spatial and structural information from the input image, they often fail to remove the suppression target when doing so requires substantial structural changes. For instance, in the second row of Figure~\ref{fig:qualitative_editing}, all editing models fail to remove the suppression target `window’ because doing so requires significant structural modification of the image. In contrast, our method generates an image that aligns with the prompt while successfully excluding the window. Additional qualitative results on FLUX-dev, FLUX-schnell, and SD3.5-Large are provided in Appendix~(Figures~\ref{fig:appendix_dev_guidance}-\ref{fig:appendix_sd35_editing}).

\begin{figure}[]
\begin{center}
   \includegraphics[width=\linewidth]{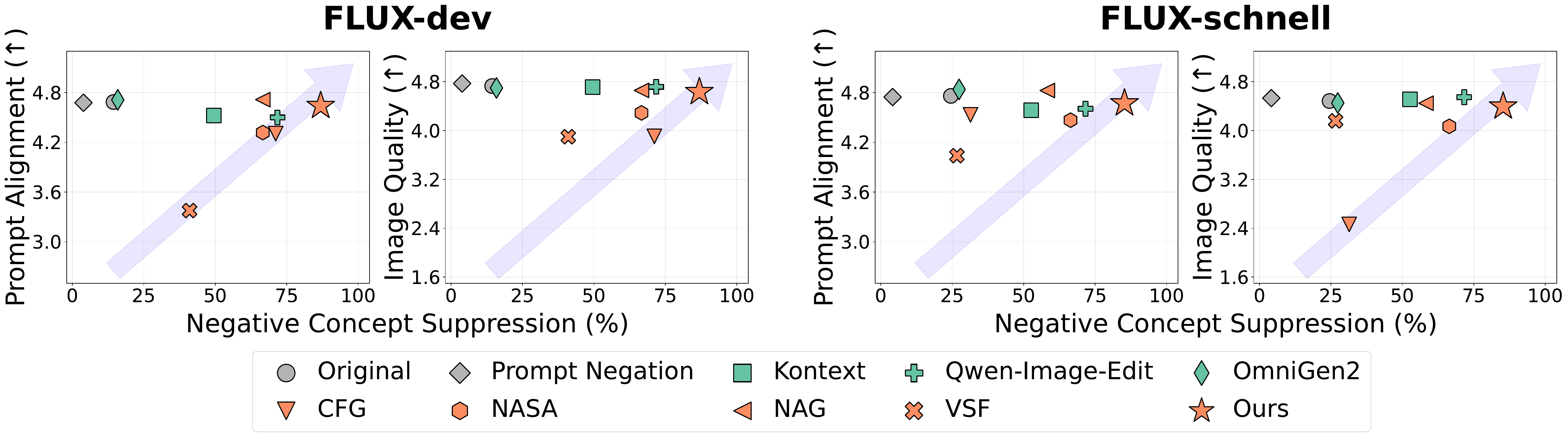}
   \vspace{-7mm}
\end{center}
\caption{
    \textbf{Quantitative comparison on FLUX-dev~(left two plots) and FLUX-schnell~(right two plots).} Each plot compares methods in Negative Concept Suppression vs. Prompt Alignment or Image Quality~(higher is better for all metrics). Gray color indicates the original model and prompt-level negation, green indicates generate-and-edit pipelines, and orange indicates negative guidance methods. Our method~($\bigstar$) outperforms other baselines in negative concept suppression while maintaining prompt alignment and image quality. Plots for each category in DCS-Bench are provided in Appendix~\ref{sec:appendix_quantitative_results}.
    \vspace{-3mm}
}

\label{fig:quantitative_comparison}
\end{figure}
\subsubsection{Quantitative Comparison.}
Figure~\ref{fig:quantitative_comparison} presents quantitative comparisons of different methods on DCS-Bench for both FLUX-dev and FLUX-schnell. For each model, we plot Negative Concept Suppression against Prompt Alignment and Image Quality, respectively. These plots also include results for a simple prompt-level negation baseline, denoted as `Prompt Negation’, which appends the phrase `without \{suppression target\}' to the prompt without using negative guidance. Detailed comparisons with this baseline are provided in Appendix~\ref{sec:appendix_prompt_negation}. For both T2I models, our method achieves the best trade-offs between Negative Concept Suppression and Prompt Alignment, as well as between Negative Concept Suppression and Image Quality, compared to all baselines. For example, on FLUX-dev, our method improves Negative Concept Suppression by at least 15\% over all other prior methods while maintaining comparable Prompt Alignment and Image Quality. These results demonstrate that our method effectively suppresses unwanted concepts while preserving the fidelity and quality of the generated images.

\vspace{-1mm}
\begin{table}[]
\centering
\setlength{\tabcolsep}{5pt}
\caption{\textbf{Results of the human preference study comparing our method with four baselines}.}
\label{tab:user_study}
\begin{tabular}{l c c c c c}
\toprule
Method & Kontext & Qwen-Image-Edit & CFG & NAG & Ours \\
\midrule
Human Preference & 36.44 & 52.03 & 36.44 & 46.22 & \textbf{70.81} \\
\bottomrule
\vspace{-10mm}
\end{tabular}
\end{table}
\vspace{-4mm}
\subsubsection{Human Preference Study.}
\label{sec:human_preference_study}
We conduct a human preference study using suppression scenarios from DCS-Bench. We select two representative methods from each baseline category, negative guidance~(CFG and NAG) and generate-and-edit pipelines~(Kontext and Qwen-Image-Edit), and compare them with our method. Participants are shown images generated by each method and asked to select all images that do not contain the suppression target while aligning well with the generation prompt. Using Amazon Mechanical Turk~(MTurk), we collected 2,220 responses from 111 valid participants. As shown in Table~\ref{tab:user_study}, our method achieves the highest human preference score, outperforming the second-best method by 18.78\%. Additional details on the human preference study, including survey questions and participant filtering criteria, are provided in Appendix~\ref{sec:appendix_human_preference_study}.

\subsection{Ablation Study}
\label{sec:ablation_study}

\begin{wrapfigure}{r}{0.36\columnwidth}
    \vspace{-10mm}
    \centering
    \includegraphics[width=\linewidth]{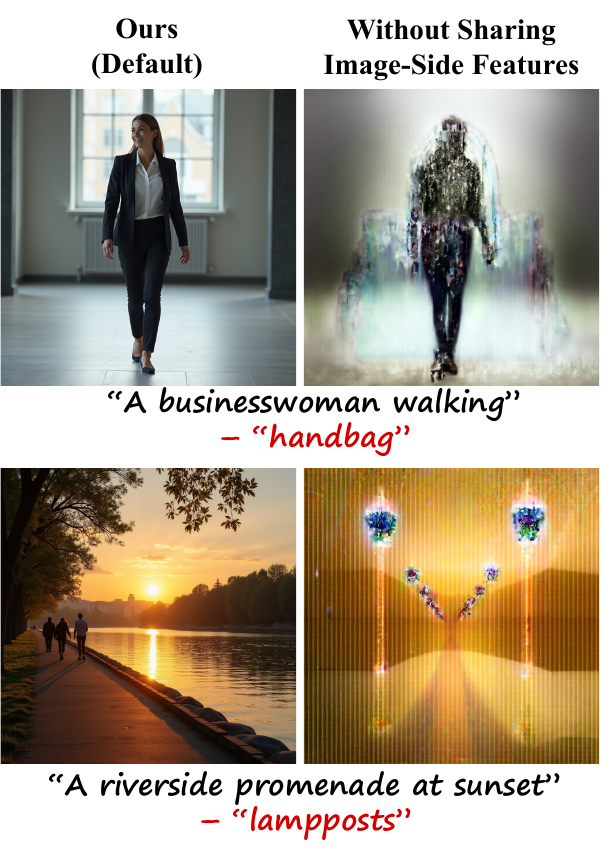}
    \vspace{-7mm}
    \caption{\textbf{Effect of image-side feature sharing.} 
    }
    \vspace{-7mm}    
    \label{fig:ablation_attention_latent_sharing}
\end{wrapfigure}
\subsubsection{Effect of Image-Side Feature Sharing.}
As explained in Eq.(\ref{eqn:negative branch}), the negative branch reuses the image-side features $(Q_{\mathrm{I}}^{+}, K_{\mathrm{I}}^{+}, V_{\mathrm{I}}^{+})$ from the positive branch while retaining its own text-side features $(Q_{\mathrm{T}}^{-}, K_{\mathrm{T}}^{-}, V_{\mathrm{T}}^{-})$. To evaluate the effect of this design, we conduct an ablation study comparing settings with and without image-side feature sharing. When the image-side features are not shared between branches, the negative branch instead uses its own image-side features $(Q_{\mathrm{I}}^{-}, K_{\mathrm{I}}^{-}, V_{\mathrm{I}}^{-})$ during the attention computation. Figure~\ref{fig:ablation_attention_latent_sharing} shows that removing feature sharing leads to spatially misaligned guidance signals, which appear as handbag-shaped and lamppost-like distortions in the first and second rows, respectively. These results indicate that sharing image-side features between branches is important for constructing spatially aligned negative guidance signals.

\subsubsection{Effect of Orthogonalization.}
\begin{figure}[b]
\centering
\begin{subfigure}[]{0.7\columnwidth}
    \centering
    \includegraphics[width=\linewidth]{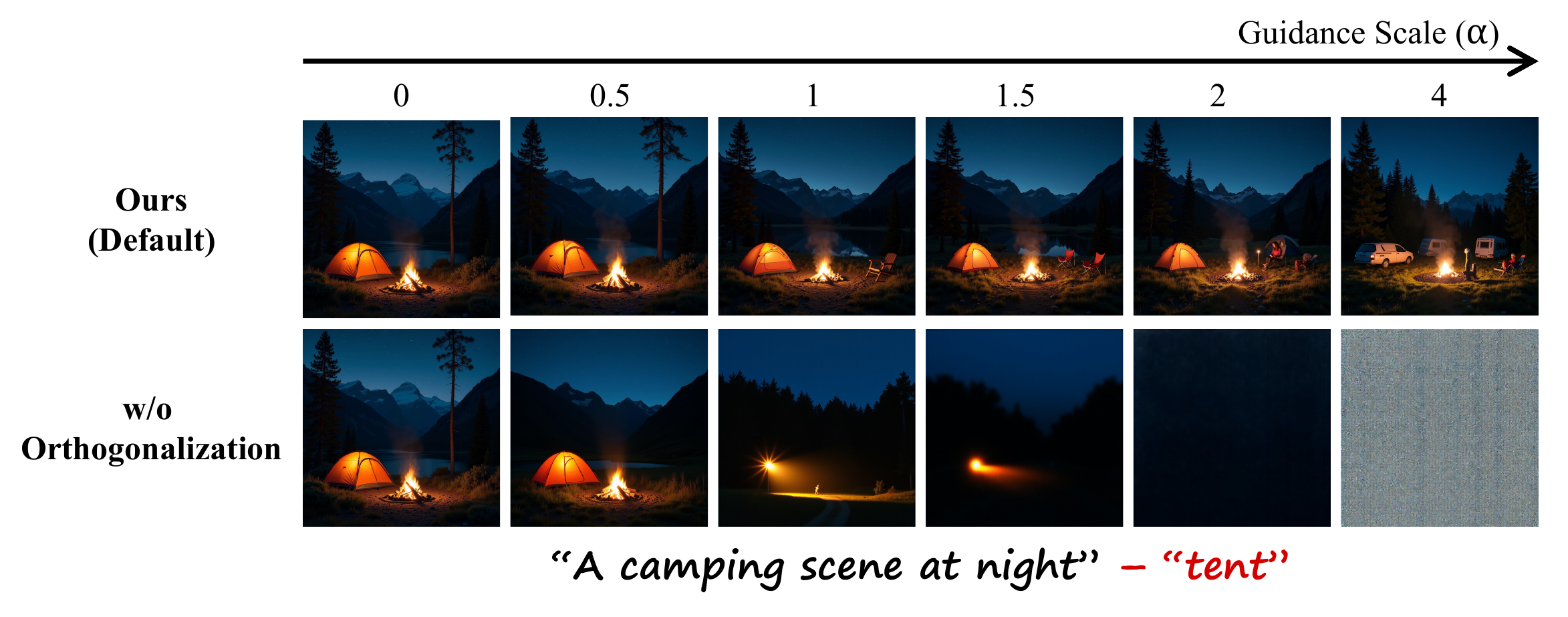}
    \caption{}
    \label{fig:ablation_orthogonal_qualitative}
\end{subfigure}
\begin{subfigure}[]{0.28\columnwidth}
    \centering
    \includegraphics[width=\linewidth]{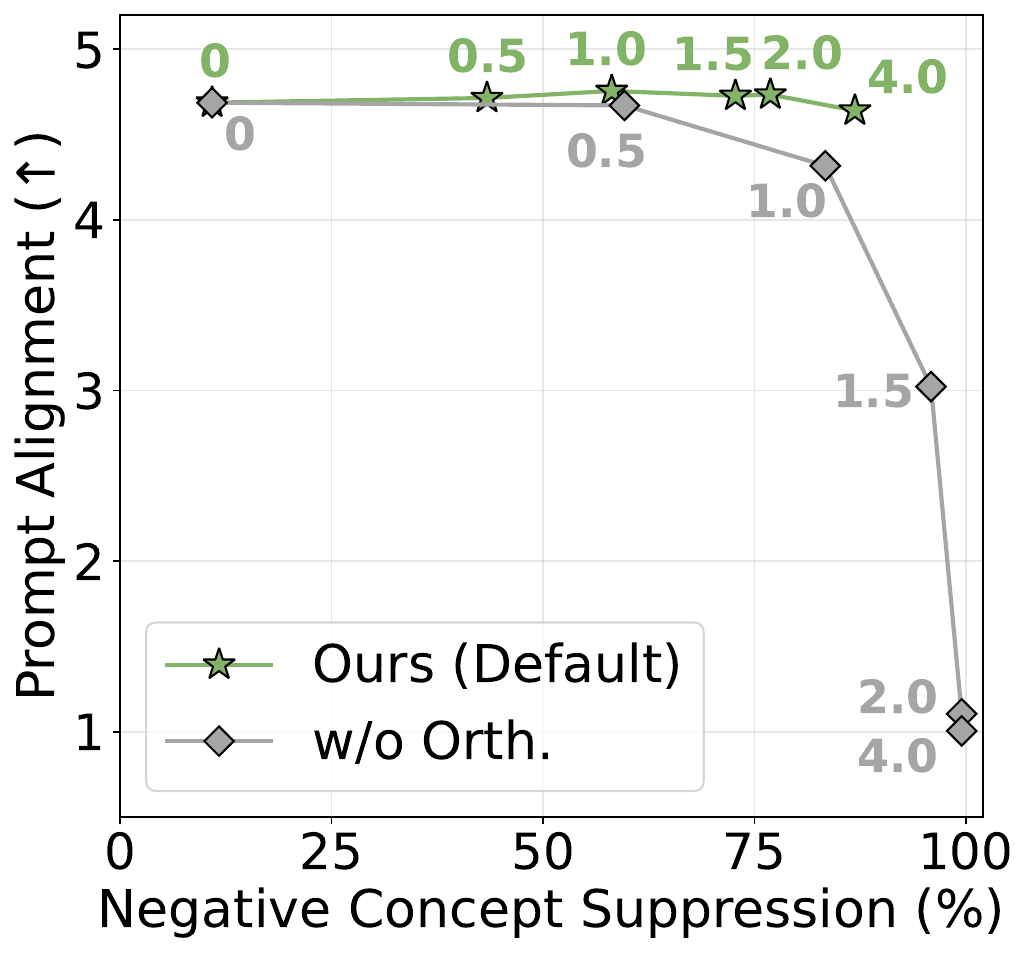}
    \caption{}
    \label{fig:ablation_orthogonal_quantitative}
\end{subfigure}
\caption{\textbf{Effect of Orthogonalization.} (a)~Qualitative comparison.
(b)~Quantitative comparison.}
\label{fig:ablation}
\end{figure}

As shown in Eq.~(\ref{eqn:Orthogonal Guidance}), our method subtracts only the component of $Z_{\mathrm{I2T}}^{-}$ that is orthogonal to $Z_{\mathrm{I2T}}^{+}$ from the positive attention feature $Z_{\mathrm{I2T}}^{+}$ to guide image generation. To evaluate the effect of this design, we compare our formulation with a variant without orthogonalization, where the entire $Z_{\mathrm{I2T}}^{-}$ is subtracted from $Z_{\mathrm{I2T}}^{+}$ during generation. Figure~\ref{fig:ablation} compares these two settings while varying the guidance scale from 0 (no negative guidance) to 4 (strong guidance). As shown in the figure, removing orthogonalization leads to severe degradation of the generated images before the target concepts are sufficiently suppressed. This behavior appears as noisy images in the bottom row of Figure~\ref{fig:ablation_orthogonal_qualitative} and extremely low Prompt Alignment scores in Figure~\ref{fig:ablation_orthogonal_quantitative}.

\section{Additional Applications}
\label{sec:additional_applications}

\begin{figure}[t]
\begin{center}
   \includegraphics[width=0.75\linewidth]{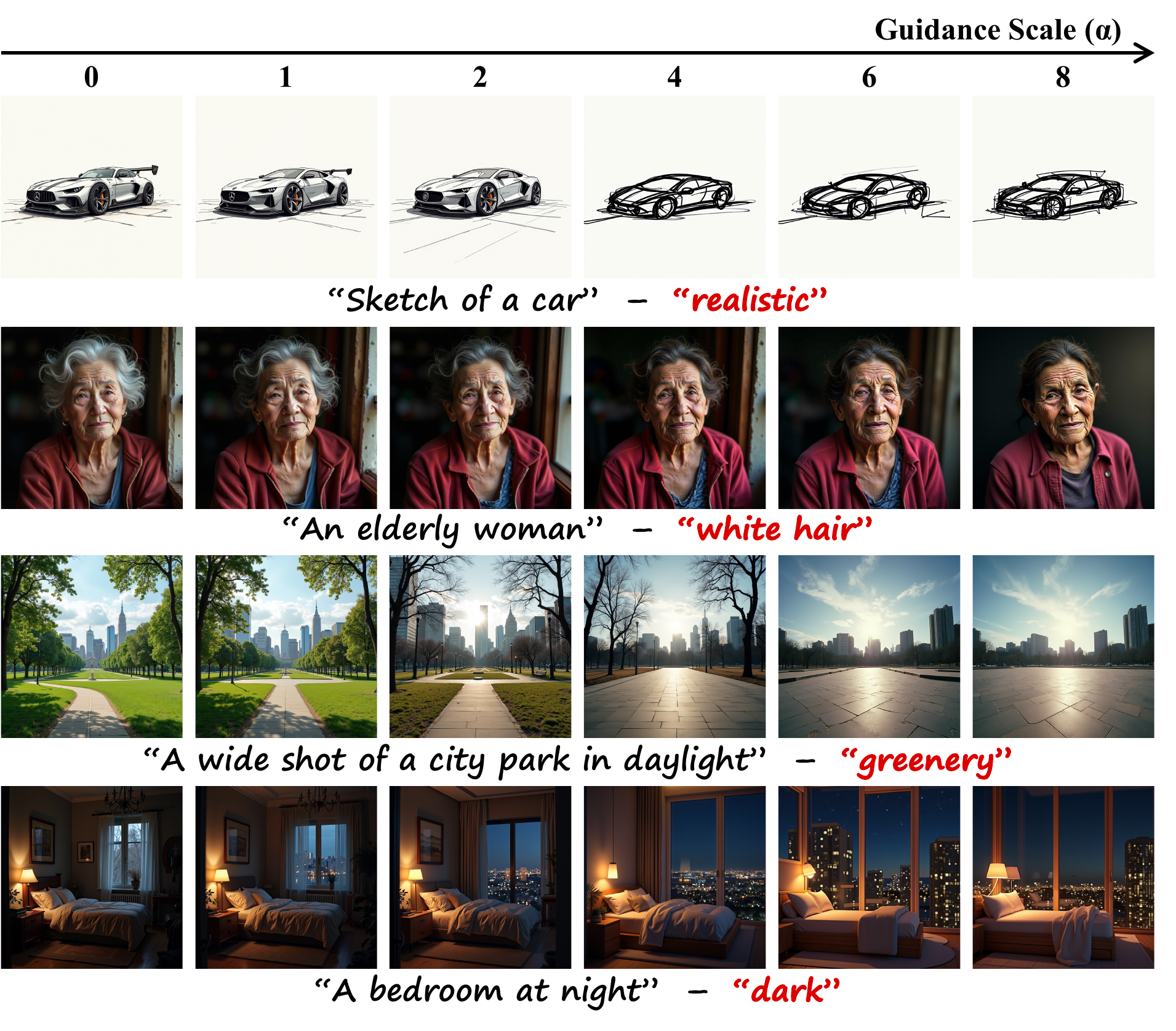}   
    \vspace{-4mm}
\end{center}
\caption{
\textbf{Examples of adjustable concept suppression.} Increasing the guidance scale $\alpha$ gradually suppresses the unwanted concept while preserving prompt alignment and image quality. $\alpha = 0$ corresponds to generation without negative guidance.
}
\vspace{-3mm}
\label{fig:scale_control} 
\end{figure}
\subsection{Adjustable Concept Suppression}
Our method also enables continuous control over suppression strength by adjusting the guidance scale $\alpha$. Figure~\ref{fig:scale_control} shows examples of this adjustable concept suppression. In the first row, increasing the guidance scale gradually shifts the generated images away from a realistic style. In the third row, increasing the guidance scale progressively reduces the greenery in the city park, eventually producing a city park with little or no greenery. Across all examples, our method maintains strong prompt alignment and image quality while adjusting the strength of suppression. Additional results are provided in Figure~\ref{fig:appendix_control_suppression} of Appendix~\ref{sec:appendix_controllable_multi_concept}.

\subsection{Multi-Concept Suppression}
\begin{figure}[t]
\begin{center}
   \includegraphics[width=0.6\linewidth]{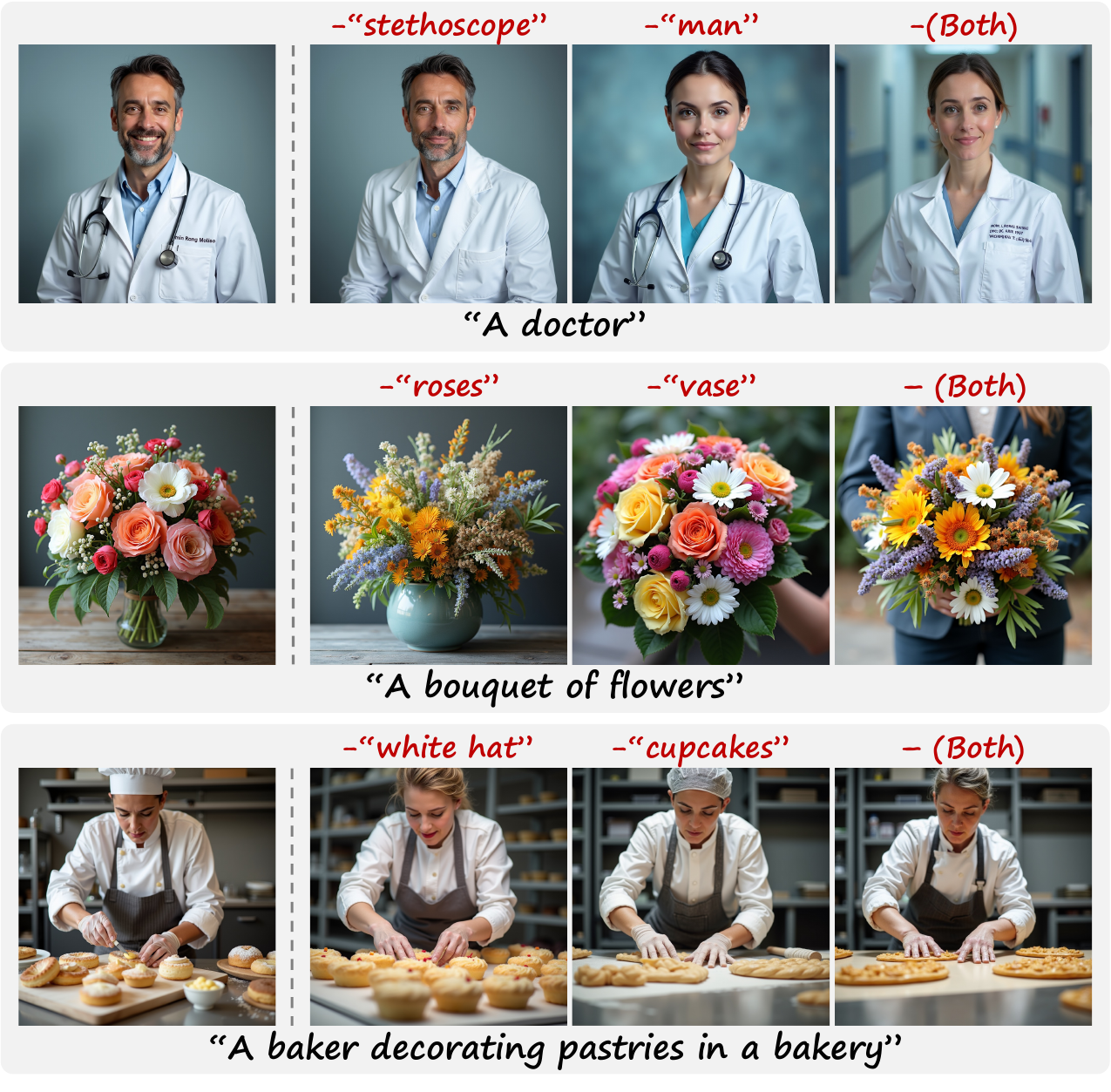}   
    \vspace{-4mm}
\end{center}
\caption{
    \textbf{Examples of multi-concept suppression.}
    Our method suppresses multiple concepts simultaneously by specifying all target concepts in a single negative prompt. The leftmost images correspond to generation without negative guidance.
}
\vspace{-3mm}
\label{fig:multi_concept}
\end{figure}

Our method can suppress multiple concepts simultaneously by specifying all target concepts in a single negative prompt. Figure~\ref{fig:multi_concept} presents several examples. In the first row, suppressing `stethoscope' and `man' individually results in the removal of the stethoscope and the replacement of the man with a woman, respectively. When both concepts are specified together in the negative prompt, these effects occur simultaneously. The remaining examples demonstrate similar behavior. Additional results are provided in Figure~\ref{fig:appendix_multi_concept} of Appendix~\ref{sec:appendix_controllable_multi_concept}.

\label{sec:conclusion}
\begin{wrapfigure}{r}{0.33\columnwidth}
    \vspace{-10mm}
    \centering
    \includegraphics[width=\linewidth]{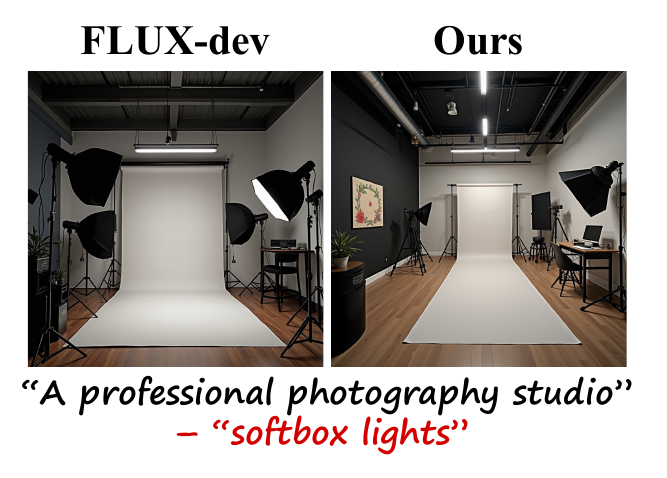}
    \vspace{-7mm}
    \caption{\textbf{Failure case.} 
    }
    \vspace{-7mm}    
    \label{fig:limitation}
\end{wrapfigure}

\section{Limitations and Conclusion}
\vspace{-2mm}
Our method has several limitations. First, because it suppresses concepts during the generation process, it is not well suited for sequential concept suppression, which is better suited to generate-and-edit pipelines. Second, our method occasionally fails to suppress certain concepts, with relatively lower suppression performance in the Place/Scene and Event/Action categories, as shown in Figure~\ref{fig:appendix_pareto_category_dev} of Appendix~\ref{sec:appendix_experimental_results}. Figure~\ref{fig:limitation} presents a representative failure case in which softbox lights remain partially visible despite suppression. These limitations could potentially be mitigated by integrating our method into the generation stage of generate-and-edit pipelines while using state-of-the-art image editing models for the editing stage.

In conclusion, we propose Orthogonal Negative Guidance in attention feature space, a new negative guidance method that effectively suppresses unwanted concepts during image generation. We also introduce the Diverse Concept Suppression Benchmark for systematic evaluation across diverse concept suppression scenarios. Extensive experiments on several rectified flow T2I models show that our method outperforms prompt-level negation, prior negative guidance methods, and generate-and-edit pipelines in suppressing unwanted concepts while maintaining prompt alignment and image quality. Finally, we demonstrate that our method supports adjustable concept suppression and simultaneous suppression of multiple concepts.

\clearpage

%
%
\bibliographystyle{splncs04}
\bibliography{bibliography}

\clearpage
\appendix
\section{Pseudocode for Orthogonal Negative Guidance}
\label{sec:appendix_method_pseudo_code}

\begin{algorithm*}[!ht]
\small
\caption{Orthogonal Negative Guidance}
\label{alg:orthogonal_negative_guidance}
\begin{algorithmic}[1]
    \REQUIRE $P^{+}$: Positive prompt
    \REQUIRE $P^{-}$: Negative prompt
    \REQUIRE $\alpha$: Guidance scale
    \REQUIRE $\tau$: Starting timestep for applying negative guidance
    \REQUIRE $N$: Total number of denoising steps
    \STATE Initialize noise latent $\mathbf{z}_0 \sim \mathcal{N}(0, I)$
    \FOR{$t = 0, 1, \dots, N-1$}
        \FOR{each MM-DiT block $l$}
            \STATE Compute $(Q_{\mathrm{T}}^{+})^{(t)}, (K_{\mathrm{T}}^{+})^{(t)}, (V_{\mathrm{T}}^{+})^{(t)}$ from the text features of $P^{+}$,
            $(Q_{\mathrm{I}}^{+})^{(t)}, (K_{\mathrm{I}}^{+})^{(t)}, (V_{\mathrm{I}}^{+})^{(t)}$ from the image features
            
            \STATE Compute $(A^{+})^{(t)} \leftarrow \mathrm{softmax}\!\left(
            \frac{[(Q_{\mathrm{T}}^{+})^{(t)};\,(Q_{\mathrm{I}}^{+})^{(t)}]
            [(K_{\mathrm{T}}^{+})^{(t)};\,(K_{\mathrm{I}}^{+})^{(t)}]^{\top}}{\sqrt{d_k}}
            \right)$
            

            \STATE Compute 
            $(Z_{\mathrm{I2T}}^{+})^{(t)} \leftarrow (A_{\mathrm{I2T}}^{+})^{(t)} (V_{\mathrm{T}}^{+})^{(t)}$

            \STATE Compute $(Z_{\mathrm{I2I}}^{+})^{(t)} \leftarrow (A_{\mathrm{I2I}}^{+})^{(t)} (V_{\mathrm{I}}^{+})^{(t)}$
            
            \STATE Set the image-modality attention output:
            $(\widehat{Z}_{\mathrm{I}}^{+})^{(t)} \leftarrow (Z_{\mathrm{I2T}}^{+})^{(t)} + (Z_{\mathrm{I2I}}^{+})^{(t)}$
            \IF{$t \geq \tau$}
                \STATE Compute $(Q_{\mathrm{T}}^{-})^{(t)}, (K_{\mathrm{T}}^{-})^{(t)}, (V_{\mathrm{T}}^{-})^{(t)}$ from the text features of $P^{-}$

                \STATE Compute 
                $(A^{-})^{(t)} \leftarrow \mathrm{softmax}\!\left(
                \frac{[(Q_{\mathrm{T}}^{-})^{(t)};\,(Q_{\mathrm{I}}^{+})^{(t)}]
                [(K_{\mathrm{T}}^{-})^{(t)};\,(K_{\mathrm{I}}^{+})^{(t)}]^{\top}}{\sqrt{d_k}}
                \right)$
                
                \STATE Compute
                $(Z_{\mathrm{I2T}}^{-})^{(t)} \leftarrow (A_{\mathrm{I2T}}^{-})^{(t)} (V_{\mathrm{T}}^{-})^{(t)}$
                
                \STATE Apply orthogonal negative guidance:
                $(\widehat{Z}_{\mathrm{I2T}}^{+})^{(t)} \leftarrow
                (Z_{\mathrm{I2T}}^{+})^{(t)} - \alpha \left[
                (Z_{\mathrm{I2T}}^{-})^{(t)} -
                \mathrm{proj}_{(Z_{\mathrm{I2T}}^{+})^{(t)}}((Z_{\mathrm{I2T}}^{-})^{(t)})
                \right]$
                
                \STATE Update 
                $(\widehat{Z}_{\mathrm{I}}^{+})^{(t)} \leftarrow
                (\widehat{Z}_{\mathrm{I2T}}^{+})^{(t)} + (Z_{\mathrm{I2I}}^{+})^{(t)}$
            \ENDIF
            \STATE Use $(\widehat{Z}_{\mathrm{I}}^{+})^{(t)}$ as the image-modality attention output for block $l$
        \ENDFOR
        \STATE Update the latent $\mathbf{z}_{t+1}$ using the model prediction
    \ENDFOR
    \STATE Decode $\mathbf{z}_N$ to obtain output image $\mathbf{x}$
    \STATE \textbf{Return:} $\mathbf{x}$
    \STATE \textit{Note:} $(A_{\mathrm{T2T}})^{(t)}, (A_{\mathrm{T2I}})^{(t)}, (A_{\mathrm{I2T}})^{(t)}, (A_{\mathrm{I2I}})^{(t)}$ denote the corresponding sub-blocks of the attention map $(A)^{(t)}$.
    \STATE \textit{Note:} $\mathrm{proj}_{Z_{\mathrm{I2T}}^{+}}(\cdot)$ denotes projection onto $Z_{\mathrm{I2T}}^{+}$ along the hidden dimension.
    \STATE \textit{Note:} $d_k$ denotes the key dimension.
    \STATE \textit{Note:} $(\cdot)^{(t)}$ denotes the features at timestep {t}.
    \STATE \textit{Note:} Text-modality attention outputs follow the standard MM-DiT computation and are omitted for clarity.
\end{algorithmic}
\end{algorithm*}

\clearpage
\section{Benchmark Construction Details}
\label{sec:appendix_benchmark}

\subsection{Dataset Construction Process}
\label{sec:appendix_benchmark_construction}
\begin{figure}[]
    \centering
    \includegraphics[width=\linewidth]{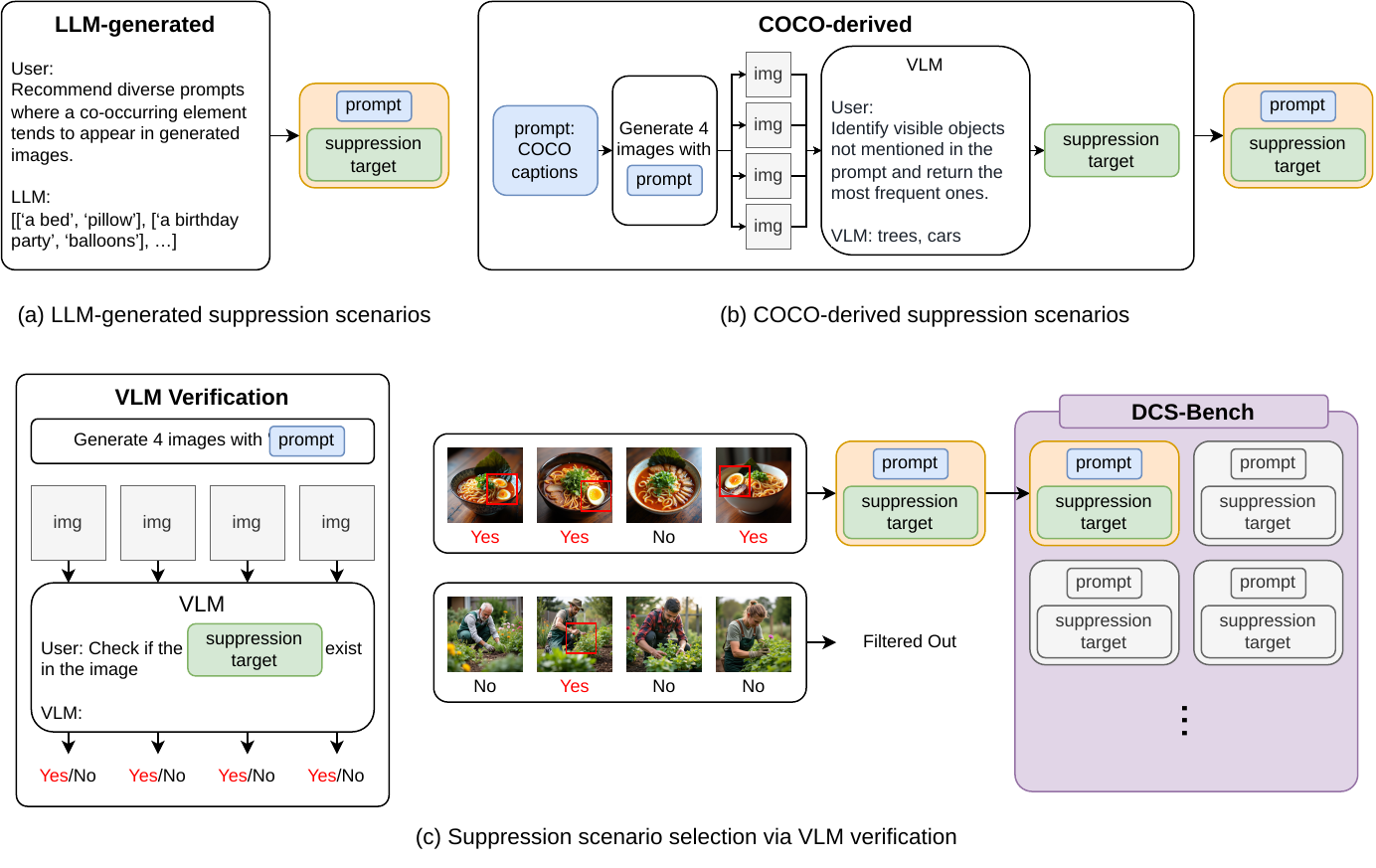}
    \vspace{-5.0mm}
    \caption{\textbf{Overview of the DCS-Bench dataset construction pipeline.} 
    (a) LLM-generated suppression scenarios.
    (b) COCO-derived suppression scenarios.
    (c) Suppression scenario selection via VLM verification, retaining only scenarios whose concept appears in at least two of four images.}
    \label{fig:benchmark_construction}
    \vspace{-7.0mm}
\end{figure}

Each concept suppression scenario in DCS-Bench pairs a generation prompt with a concept that FLUX-dev~\cite{flux2024} frequently generates despite not being explicitly specified in the prompt. We construct these scenarios from two complementary sources.
First, an LLM generates candidate (\{prompt\}, \{suppression target\}) pairs. 
Second, COCO~\cite{lin2014microsoft} captions are used as prompts, and a VLM identifies concepts that appear in the generated images but are not specified in the caption. 
Figure~\ref{fig:benchmark_construction} illustrates the construction pipeline. The final benchmark contains 200 validated scenarios.

\subsubsection{LLM-generated scenarios.}
We prompt an LLM (GPT-4o~\cite{hurst2024gpt}) to generate candidate scenarios in batches.
Each batch is filtered using the LLM: concepts essential to the scene's identity or logical validity are removed, retaining only those that are commonly associated but not strictly required.
This process is repeated until 100 verified scenarios are collected, screening approximately 600 candidates in total.

\subsubsection{COCO-derived scenarios.}
We sample image captions from COCO~\cite{lin2014microsoft} as prompts in batches, balancing the category distribution to promote diversity.
For each caption, we generate four images using the T2I model.
We then prompt a VLM (GPT-4o-mini~\cite{hurst2024gpt}) with the four images and the caption to identify visible objects not mentioned in the caption, returning up to three candidates.
Among these, the most frequent object is selected as the suppression target.
This process is repeated until 100 verified scenarios are collected, screening approximately 500 captions in total.

\subsubsection{Verification.}
To ensure that each association is not incidental, we generate four images per prompt using the T2I model under different random seeds and retain only pairs in which the target concept appears in at least two of the four images.
Concept presence is assessed by the VLM.
This threshold balances coverage and reliability: a lower threshold would admit incidental pairings, while a stricter one would exclude pairs with meaningful associations.

\subsection{6 Categories in DCS-Bench}
\label{sec:appendix_benchmark_types}
Each category is defined by the semantic relationship between the prompt and the suppression target:
\begin{itemize}
    \item \textbf{Associated Concept for Place/Scene} (77 scenarios). The prompt describes a place or scene, and the suppression target is a concept commonly associated with that setting, which T2I models frequently generate even when it is not mentioned in the prompt (e.g., bathroom $\to$ bathtub, playground $\to$ slide).
    \item \textbf{Associated Concept for Event/Action} (47 scenarios). The prompt describes an event or activity, and the suppression target is a concept commonly associated with it (e.g., birthday party $\to$ balloons, camping $\to$ tent).
    \item \textbf{Co-occurring Object for an Object} (29 scenarios). The prompt describes an object, and the suppression target is another independently existing object that frequently co-occurs with it (e.g., desk $\to$ chair, piano $\to$ sheet music).
    \item \textbf{Dominant Subtype of a Supercategory} (19 scenarios). The prompt describes a supercategory in a specific context, and the suppression target is its subtype that T2I models frequently generate under that context. We refer to such a subtype as a \textit{dominant subtype}, meaning that it tends to dominate generation in the context specified by the prompt rather than in general. For example, for the prompt `A bunch of animals that are in a field', T2I models often generate deer instead of other animals, indicating that deer is the dominant subtype of `animals' in this context.
    \item \textbf{Associated Component of an Object} (18 scenarios). The prompt describes an object, and the suppression target is a constituent part or attached element commonly generated with it (e.g., bowl of ramen $\to$ boiled egg, branch $\to$ leaves).
    \item \textbf{Associated Concept for Occupation/Role} (10 scenarios). The prompt describes a professional role, and the suppression target is a concept commonly associated with that role (e.g., doctor $\to$ stethoscope, magician $\to$ top hat).
\end{itemize}

The full list of all 200 scenarios grouped by category is provided in Appendix~\ref{sec:appendix_full_pair_list}.

\subsection{Generalization of DCS-Bench Across T2I Models}
\label{sec:appendix_cross_model}
To construct natural suppression targets for evaluating concept suppression performance, we build DCS-Bench by collecting pairs of prompts and their frequently co-occurring concepts. Because these pairs are identified using FLUX-dev, they may not necessarily constitute valid suppression scenarios for other T2I models such as FLUX-schnell~\cite{flux2024} or SD3.5-Large~\cite{sd3.5}. To verify their validity for these models, we examine whether the suppression targets identified by FLUX-dev also frequently appear in images generated by these models when the concepts are not explicitly mentioned in the prompts. Table~\ref{tab:cross_model_co_occurrence} summarizes the results. Both FLUX-schnell and SD3.5-Large exhibit frequent co-occurrence in more than 75\% of the 200 suppression scenarios in DCS-Bench, indicating that the benchmark remains effective for evaluating concept suppression across different T2I models.

\begin{table}[h]
\centering
\caption{\textbf{Cross-model co-occurrence rates on the 200 benchmark pairs.} Each pair is considered co-occurring if the associated concept appears in at least 2 of 4 generated images.}
\label{tab:cross_model_co_occurrence}
\begin{tabular}{lc}
\toprule
Model & Co-occurrence Rate (\%) \\
\midrule
FLUX-dev$^\dagger$ & 100.0 \\
FLUX-schnell & 83.0 \\
SD3.5-Large & 75.5 \\
\bottomrule
\end{tabular}

\vspace{0.3em}
\parbox{0.96\linewidth}{\footnotesize
$^\dagger$ Used for benchmark construction.}
\end{table}

\clearpage
\section{Experimental Details}
\label{sec:appendix_experimental_details}

\subsection{Implementation Details}
\label{sec:appendix_implementation_details}

We apply our method to several rectified flow text-to-image (T2I) models: FLUX-dev~\cite{flux2024}, FLUX-schnell~\cite{flux2024}, and SD3.5-Large~\cite{sd3.5}. 
Within each model, it is applied to all MM-DiT blocks at timesteps $t \geq \tau$. 
For FLUX-dev and SD3.5-Large, we use 28 sampling steps with guidance scale $\alpha = 4$ starting from timestep $\tau = 2$. 
For FLUX-schnell, we use 4 sampling steps with $\alpha = 2$ from $\tau = 0$. 
For SD3.5-Large, we additionally apply negative guidance using the image-to-text attention output computed up to the EOT token. 
All experiments are conducted on a single NVIDIA A6000 GPU.

All negative guidance baselines and the prompt-level negation baseline use the same sampling configuration as the corresponding backbone: 28 steps for FLUX-dev and SD3.5-Large, and 4 steps for FLUX-schnell. 
For prompt-level negation, we append the phrase `without \{suppression target\}' to the prompt. 
For CFG~\cite{ho2022classifier}, we use a guidance scale of 3.5 for FLUX-dev and FLUX-schnell, and 7 for SD3.5-Large. 
For NASA~\cite{nguyen2025supercharged}, we use the reimplementation provided in the official VSF~\cite{guo2025vsf} GitHub repository, as the original code has not been publicly released, and set the guidance scale to 0.2 for FLUX-dev and 0.25 for FLUX-schnell based on a small hyperparameter sweep. 
For NAG~\cite{chen2025normalized}, we follow the official implementation and set the NAG scale to 4 for all three models. 
For VSF~\cite{guo2025vsf}, we follow the official implementation and set the guidance scale to 8 for FLUX-dev and 6 for FLUX-schnell. 
For both NASA and VSF, we report results on FLUX-dev and FLUX-schnell only, as we were unable to find hyperparameters that produced satisfactory results on SD3.5-Large.

For generate-and-edit baselines, images are first generated by the corresponding backbone and then edited using the instruction `Remove \{concept\}.' 
For Kontext~\cite{labs2025flux} and OmniGen2~\cite{wu2025omnigen2}, we use their respective default settings. 
For Qwen-Image-Edit~\cite{wu2025qwenimagetechnicalreport}, we use the quantized model due to GPU memory constraints, with a guidance scale of 4 and 28 inference steps.

\subsection{Evaluation Details}
\label{sec:appendix_evaluation}

Figures~\ref{fig:neg_invisibility_score_prompt}--\ref{fig:image_quality_prompt} show the prompt templates used for the three evaluation metrics: Negative Concept Suppression, Prompt Alignment, and Image Quality.

To validate the reliability of the VLM-based evaluation, we compared VLM scores with human annotations.
We randomly sampled 15 images per method (120 images in total) and asked three annotators to independently rate each image on all three metrics. 
The final human rating for each image was computed as the average of the three annotations. 
As shown in Table~\ref{tab:vlm_human_correlation}, the VLM achieves 91\% agreement for Negative Concept Suppression and strong Pearson correlations~($r \geq 0.71$) for Prompt Alignment and Image Quality, indicating that the VLM-based metrics align well with human judgments.

\begin{table}[h]
\centering
\setlength{\tabcolsep}{8pt}
\caption{\textbf{VLM-human agreement per image ($N = 120$).} For Negative Concept Suppression (binary), we measure agreement as classification accuracy. For Prompt Alignment and Image Quality (1--5 scale), we report Pearson correlation ($p < 0.001$).}
\label{tab:vlm_human_correlation}
\begin{tabular}{l c c}
\toprule
Metric & Measure & Value \\
\midrule
Negative Concept Suppression & Accuracy & 0.91 \\
Prompt Alignment & Pearson $r$ & 0.71 \\
Image Quality & Pearson $r$ & 0.77 \\
\bottomrule
\end{tabular}
\end{table}

\begin{figure}[h]
\centering
\includegraphics[width=0.95\linewidth]{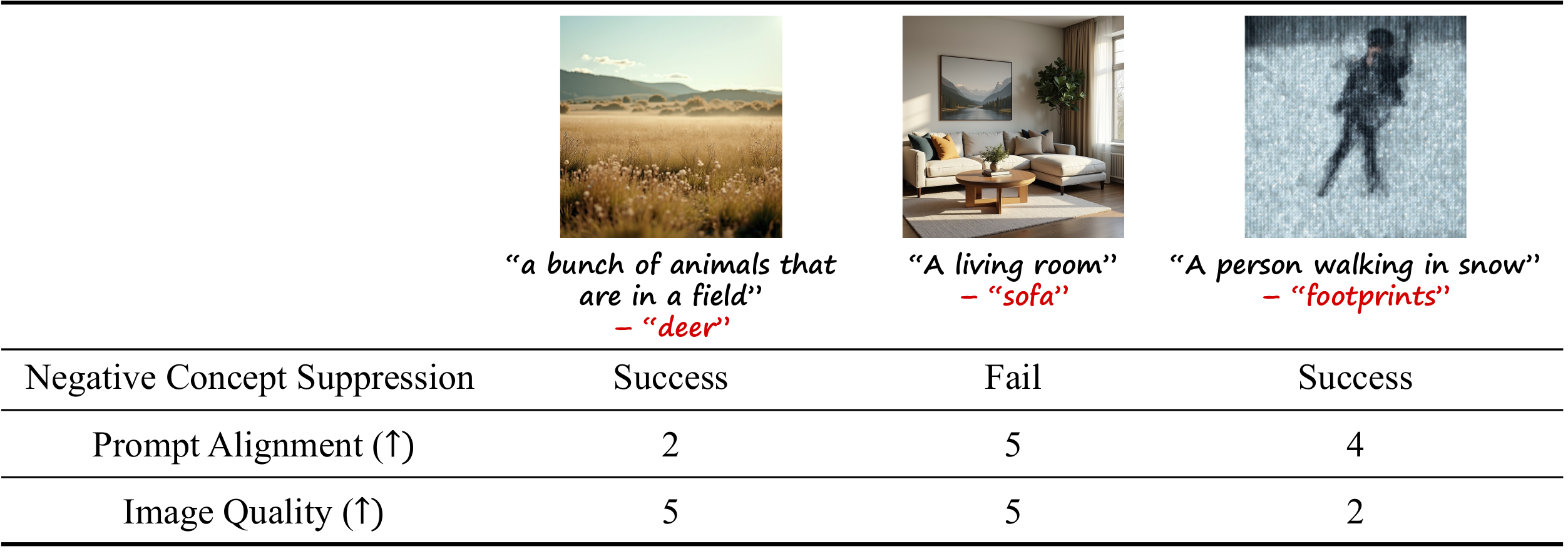}
\caption{\textbf{Representative examples from the VLM-based evaluation across the three metrics.} 
Negative Concept Suppression indicates whether the target concept is absent from the generated image. 
Prompt Alignment and Image Quality are rated by the VLM on a 1--5 scale.}
\vspace{-1mm}
\label{fig:vlm_score_examples}
\end{figure}

Figure~\ref{fig:vlm_score_examples} shows representative examples from the VLM-based evaluation across the three metrics. 
The examples suggest that the VLM assigns scores that are generally consistent with human perception. 
The figure also illustrates that the three metrics capture different aspects of the generated images; for example, an image may achieve successful suppression while receiving a low image quality score.

\begin{figure}[]
    \centering
    \begin{tcolorbox}[colback=gray!5, colframe=gray!50, title={Prompt Template for Negative Concept Suppression}]
        \small\ttfamily
        You are a judge evaluating whether \{suppression target\} is absent from a given image (i.e., not visible and not identifiable). \\
        You must provide exactly one binary label indicating whether the concept is absent from the image. \\
        Label definitions (Yes/No): \\
        - Yes (Absent): \\
        \hspace{1em}\{suppression target\} is not visible, cannot be identified with confidence, or is entirely absent from the image. \\
        - No (Not absent): \\
        \hspace{1em}\{Suppression target\} is clearly visible and can be unambiguously identified in the image. \\
        Important: Base the label only on whether the concept is visually observable and identifiable in the image. \\
        Output format: Return Yes or No only. Do not provide any explanations.
    \end{tcolorbox}
    \vspace{-3.0mm}
    \caption{\textbf{Prompt template for the Negative Concept Suppression metric.}
    The VLM assigns a binary label (Yes/No) indicating whether the suppression target is absent from the image.
    The placeholder \texttt{\{suppression target\}} is dynamically populated for each evaluation instance.}
    \label{fig:neg_invisibility_score_prompt}
\end{figure}

\begin{figure}[]
    \centering
    \begin{tcolorbox}[colback=gray!5, colframe=gray!50, title={Prompt Template for Prompt Alignment Scoring}]
        \small\ttfamily
        You are a judge evaluating how well a given image aligns with a given text prompt. \\
        You must provide exactly one score: a Prompt Alignment Score (1--5) indicating how well the semantic content of the image matches the content specified in the prompt. \\
        Scoring criteria: \\
        - 5: Perfect alignment: \\
        \hspace{1em}The image fully and accurately reflects all essential elements of the prompt, including objects, attributes, actions, and relationships, with no noticeable inconsistencies. \\
        - 4: Strong alignment: \\
        \hspace{1em}The image closely follows the prompt, with only minor omissions or inaccuracies that do not significantly affect the overall intent. \\
        - 3: Moderate alignment: \\
        \hspace{1em}The image captures the main idea of the prompt but contains noticeable missing elements, inaccuracies, or ambiguities. \\
        - 2: Weak alignment: \\
        \hspace{1em}The image reflects some aspects of the prompt but misses or misrepresents many important elements. \\
        - 1: Very poor alignment: \\
        \hspace{1em}The image largely fails to reflect the prompt; most key elements are missing, incorrect, or contradictory. \\
        Important: The score should be based solely on semantic alignment between the image and the prompt. \\
        Input text prompt: \{prompt\} \\
        Output format: Return a single integer from 1 to 5. Do not provide any explanations.
    \end{tcolorbox}
    \vspace{-3.0mm}
    \caption{\textbf{Prompt template for the Prompt Alignment metric.}
    The VLM assigns a score from 1 to 5 indicating how well the image aligns with the input prompt.
    The placeholder \texttt{\{prompt\}} is dynamically populated for each evaluation instance.}
    \label{fig:prompt_alignment_prompt}
    \vspace{-5.0mm}
\end{figure}

\begin{figure}[]
    \centering
    \begin{tcolorbox}[colback=gray!5, colframe=gray!50, title={Prompt Template for Image Quality Scoring}]
        \small\ttfamily
        You are a judge evaluating the visual quality of a given image, with a particular focus on visual artifacts and rendering defects. \\
        You must provide exactly one score: an Image Quality Score (1--5) indicating how free the image is from visual artifacts and rendering issues. \\
        Scoring criteria: \\
        - 5: Excellent quality: \\
        \hspace{1em}The image contains no noticeable visual artifacts. Rendering is clean and coherent, with correct structure, proportions, lighting, and textures. \\
        - 4: Good quality: \\
        \hspace{1em}The image is largely free of artifacts, with only minor imperfections that do not significantly affect visual coherence or interpretability. \\
        - 3: Moderate quality: \\
        \hspace{1em}The image contains noticeable artifacts or imperfections (e.g., mild distortion, blurring, or inconsistencies) that somewhat degrade visual quality. \\
        - 2: Poor quality: \\
        \hspace{1em}The image exhibits significant visual artifacts, such as distortions, malformed or duplicated structures, unnatural textures, or severe blurring. \\
        - 1: Very poor quality: \\
        \hspace{1em}The image is blank, nearly blank, or consists of a uniform color, indicating a rendering failure or missing visual content. \\
        Important: The score should be based solely on visual quality and the presence of artifacts, not on semantic alignment with any text prompt. \\
        Output format: Return a single integer from 1 to 5. Do not provide any explanations.
    \end{tcolorbox}
    \vspace{-3.0mm}
    \caption{\textbf{Prompt template for the Image Quality metric.}
    The VLM assigns a score from 1 to 5 indicating the visual quality of the image, focusing on artifacts and rendering defects.}
    \label{fig:image_quality_prompt}
    \vspace{-5.0mm}
\end{figure}

\clearpage

\subsection{Editing Instructions for Generate-and-Edit Pipelines}
\label{sec:appendix_editing_instruction}

In our main experiments, we use the editing instruction `Remove \{suppression target\}.' for the editing models in generate-and-edit pipelines. In this section, we also evaluate four alternative editing instructions to assess the effect of different editing instructions used in this pipeline. For this experiment, we use FLUX-dev for the generation stage and Qwen-Image-Edit for the editing stage of the generate-and-edit pipeline. As shown in Table~\ref{tab:editing_instruction}, all generate-and-edit variants yield similar results across all three metrics, indicating that different editing instructions do not substantially change the suppression performance of the generate-and-edit pipeline. These results suggest that tuning editing instructions for generate-and-edit pipelines provides only limited improvement, as all variants show substantially inferior suppression performance compared to our Orthogonal Negative Guidance.

\vspace{-3mm}
\begin{table}[h]
\centering
\caption{\textbf{Editing instructions evaluated in the ablation study.}}
\label{tab:editing_instruction_defs}
\begin{tabular}{lp{0.75\linewidth}}
\toprule
Editing Instruction & Template\\
\midrule
Instr. 1~(Default) & `Remove \{concept\}.' \\
Instr. 2 & `Remove \{concept\} while keeping the image consistent with the description: \{generation prompt\}.' \\
Instr. 3 & `Edit the image to remove \{concept\}.' \\
Instr. 4 & `Erase \{concept\} from the image.' \\
Instr. 5 & `Edit the image to remove \{concept\} while preserving the original scene described by: \{generation prompt\}.' \\
\bottomrule
\end{tabular}
\end{table}

\begin{table}[h]
\centering
\caption{\textbf{Ablation of editing instructions used in generate-and-edit pipelines.}
The instruction templates are defined in Table~\ref{tab:editing_instruction_defs}. 
Higher is better for all metrics.}
\label{tab:editing_instruction}
\setlength{\tabcolsep}{3pt}
\begin{tabular}{lccccc|c}
\toprule
Metric & Instr. 1 & Instr. 2 & Instr. 3 & Instr. 4 & Instr. 5 & Ours\\
\midrule
Negative Concept Suppression & 71.75 & 67.25 & 70.88 & 70.12 & 68.38 & 86.88\\
Prompt Alignment & 4.50 & 4.58 & 4.50 & 4.48 & 4.59 & 4.64\\
Image Quality & 4.72 & 4.73 & 4.72 & 4.72 & 4.72 & 4.63\\
\bottomrule
\end{tabular}
\end{table}

\vspace{-3mm}

\subsection{Human Preference Study}
\label{sec:appendix_human_preference_study}

\begin{figure}[]
    \vspace{-5mm}
    \centering
    \includegraphics[width=0.45\linewidth]{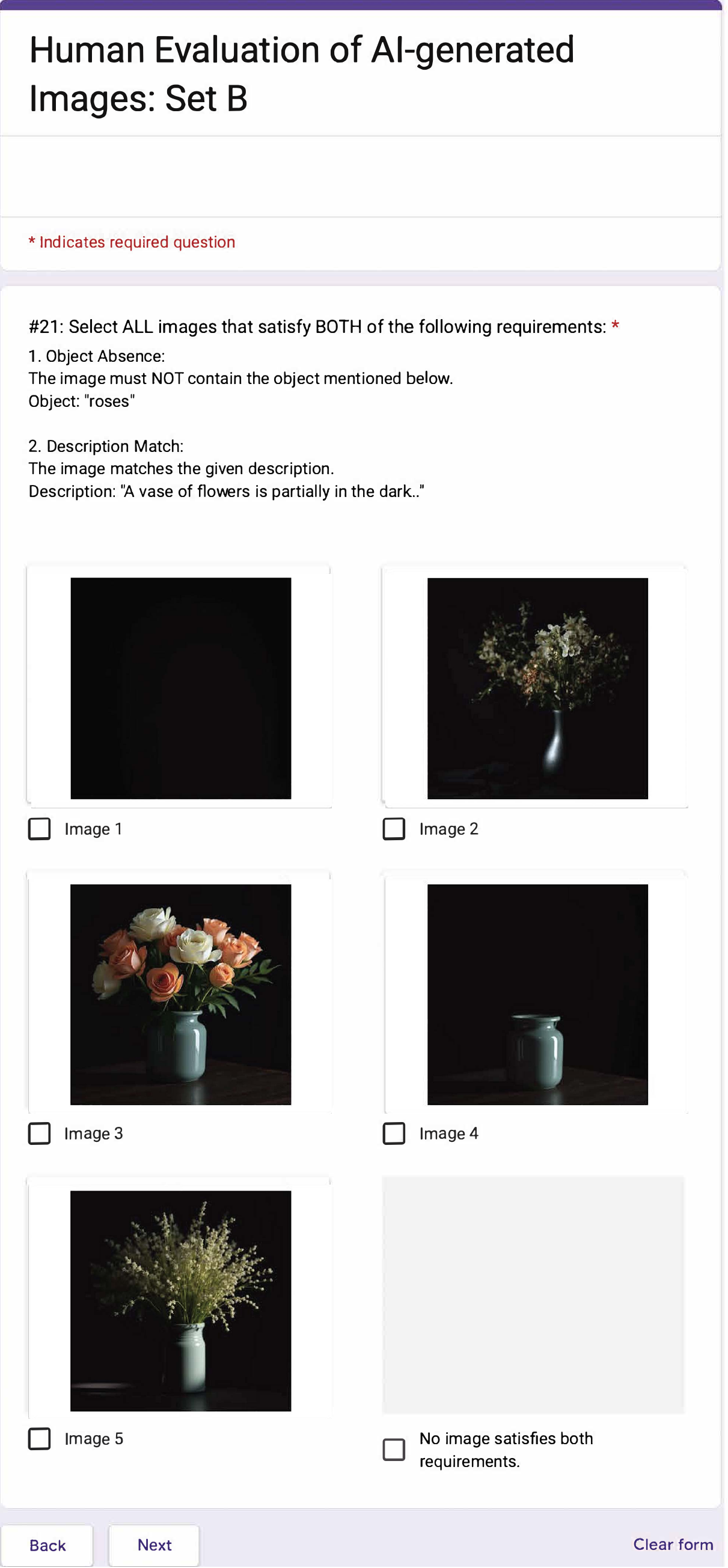}
    \caption{\textbf{Screenshot of the survey interface used in the human preference study.} Participants selected all images that satisfied both the prompt description and the object-absence constraint.}
    \label{fig:human_study_screenshot}
\end{figure}

In the human preference study, we compared five methods: two negative guidance methods (CFG and NAG), two generate-and-edit pipelines (Kontext and Qwen-Image-Edit), and our method.
We randomly sampled 100 suppression scenarios from DCS-Bench and collected responses through Amazon Mechanical Turk~(MTurk), requiring participants to have over 500 HIT approvals, an approval rate of 98\%, and US residency.
For each suppression scenario, participants were shown five images generated by the compared methods and asked to select \emph{all} images in which the unwanted concept was absent and the image matched the given prompt.
Selecting none was also permitted.
Each survey contains 20 pairs plus one dummy question to filter inattentive responses.
The correct answer to the dummy question was provided in the survey instructions, and only participants who answered it correctly were included in the final set.
This yielded 111 valid participants and 2,220 responses in total.
A screenshot of the survey interface is shown in Figure~\ref{fig:human_study_screenshot}.

\subsection{Computational Cost}
\label{sec:appendix_computational_cost}
Table~\ref{tab:appendix_computational_cost} reports the average inference time per image on a single NVIDIA A6000 GPU.
The inference time of our method is 46.73s per image, which is comparable to NASA (47.88s) and NAG (45.02s), also remaining substantially faster than
generate-and-edit pipelines (122.84s on average).

\begin{table}[]
\centering
\small
\caption{\textbf{Average inference time per generated image~(in seconds).}}
\label{tab:appendix_computational_cost}
\resizebox{\linewidth}{!}{%
\begin{tabular}{lccccccccc}
\toprule
& & \multicolumn{5}{c}{Negative Guidance} & \multicolumn{3}{c}{Generate-and-Edit Pipelines} \\
\cmidrule(lr){3-7} \cmidrule(lr){8-10}
Method & FLUX-dev & CFG & NASA & NAG & VSF & Ours & Kontext & Qwen-Image-Edit & OmniGen2 \\
\midrule
Inference Time (s) $\downarrow$ & 25.59 & 50.39 & 47.88 & 45.02 & 29.45 & 46.73 & 78.52 & 153.91 & 136.10 \\
\bottomrule
\end{tabular}
}

\vspace{0.3em}
\end{table}

\clearpage
\section{Additional Experimental Results}
\label{sec:appendix_experimental_results}
\subsection{Additional Qualitative Results}
\label{sec:appendix_qualitative_results}

\begin{figure*}[t]
\begin{center}
   \includegraphics[width=0.9\linewidth]{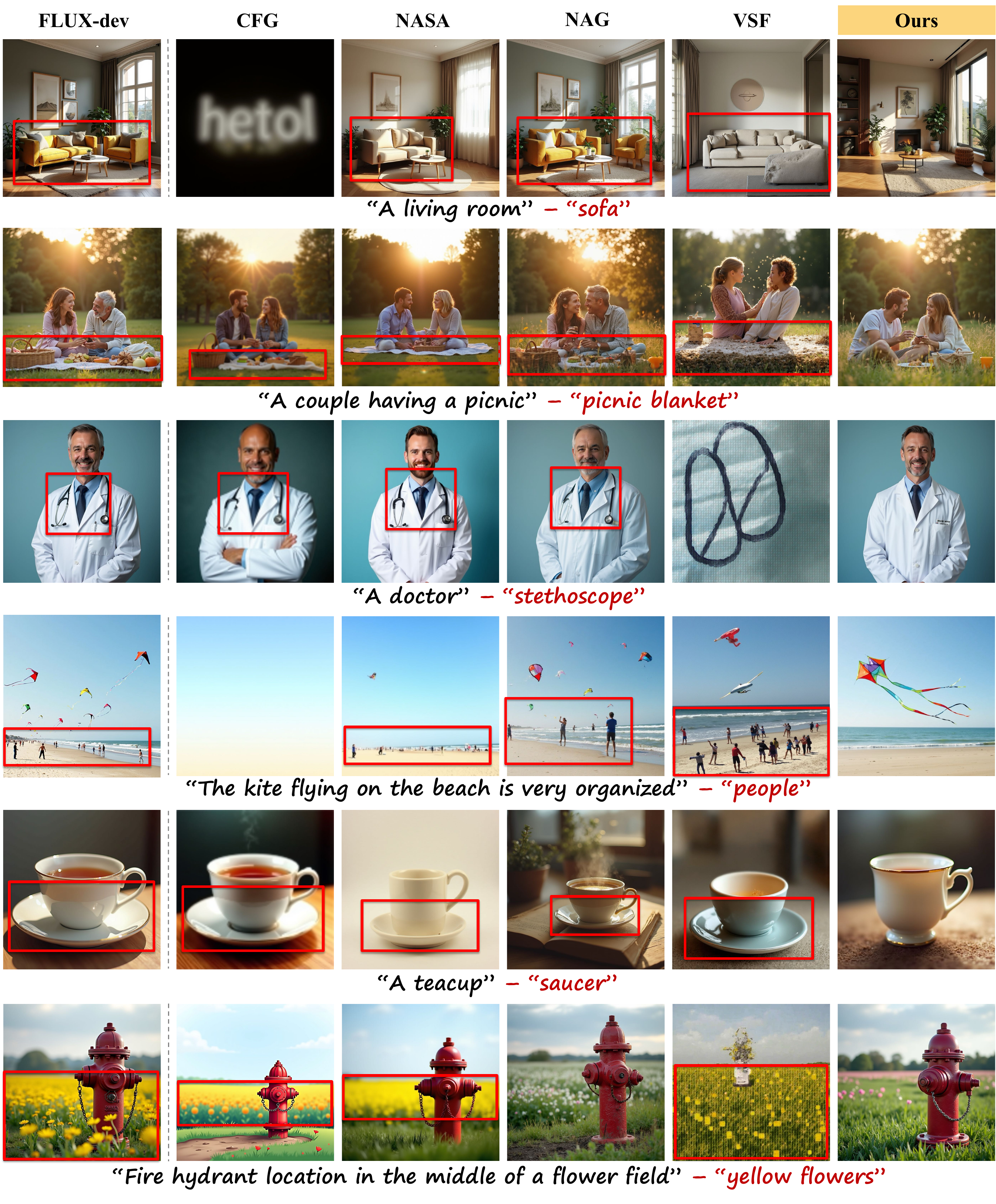}
    \vspace{-2mm}
\end{center}
\caption{\textbf{Additional qualitative comparison with negative guidance methods on FLUX-dev.}}
\vspace{-1mm}
\label{fig:appendix_dev_guidance}
\end{figure*}

\begin{figure*}[t]
\begin{center}
   \includegraphics[width=0.85\linewidth]{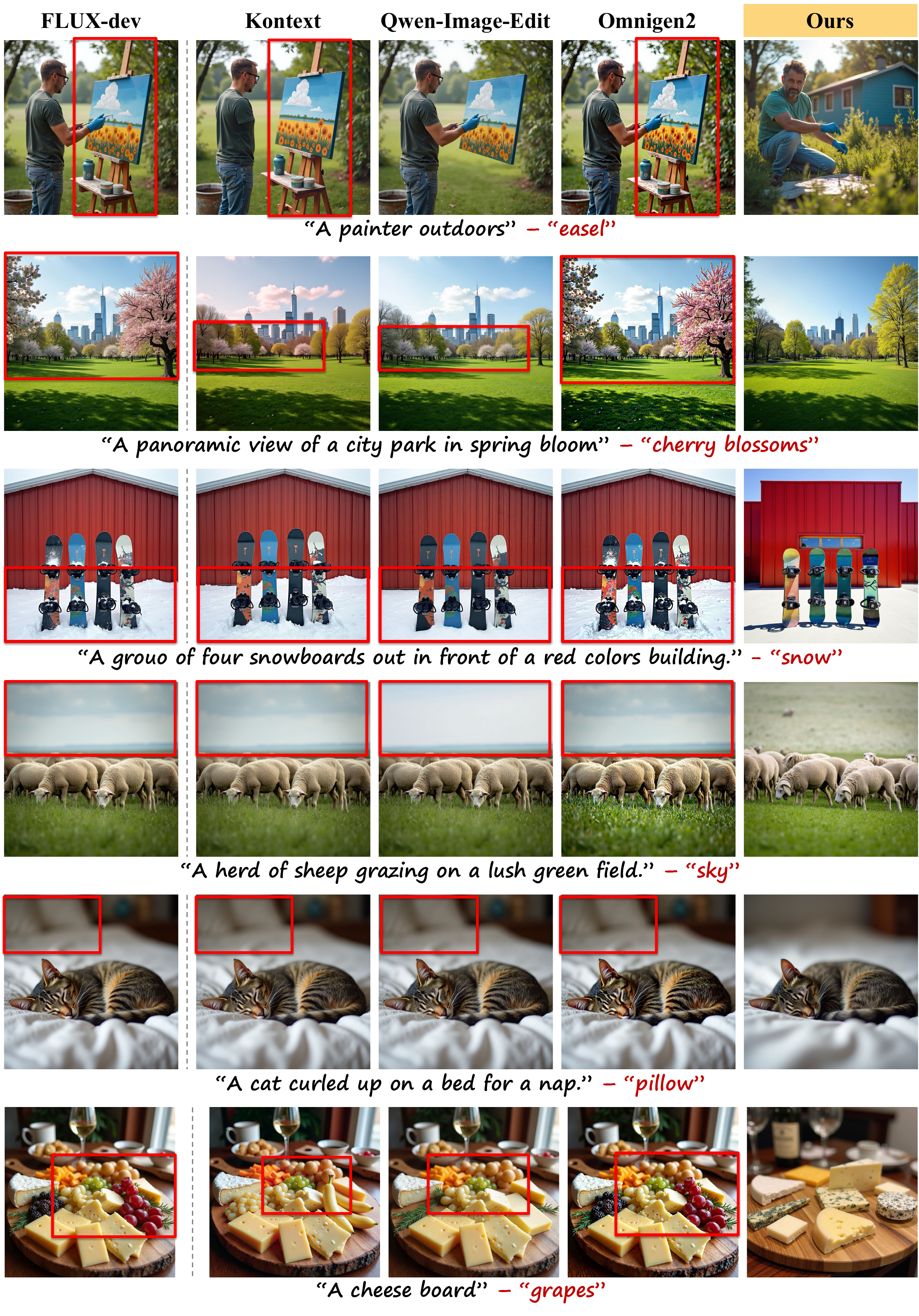}
    \vspace{-2mm}
\end{center}
\caption{\textbf{Additional qualitative comparison with generate-and-edit baselines on FLUX-dev.}}
\vspace{-1mm}
\label{fig:appendix_dev_editing}
\end{figure*}

\begin{figure*}[t]
\begin{center}
   \includegraphics[width=0.9\linewidth]{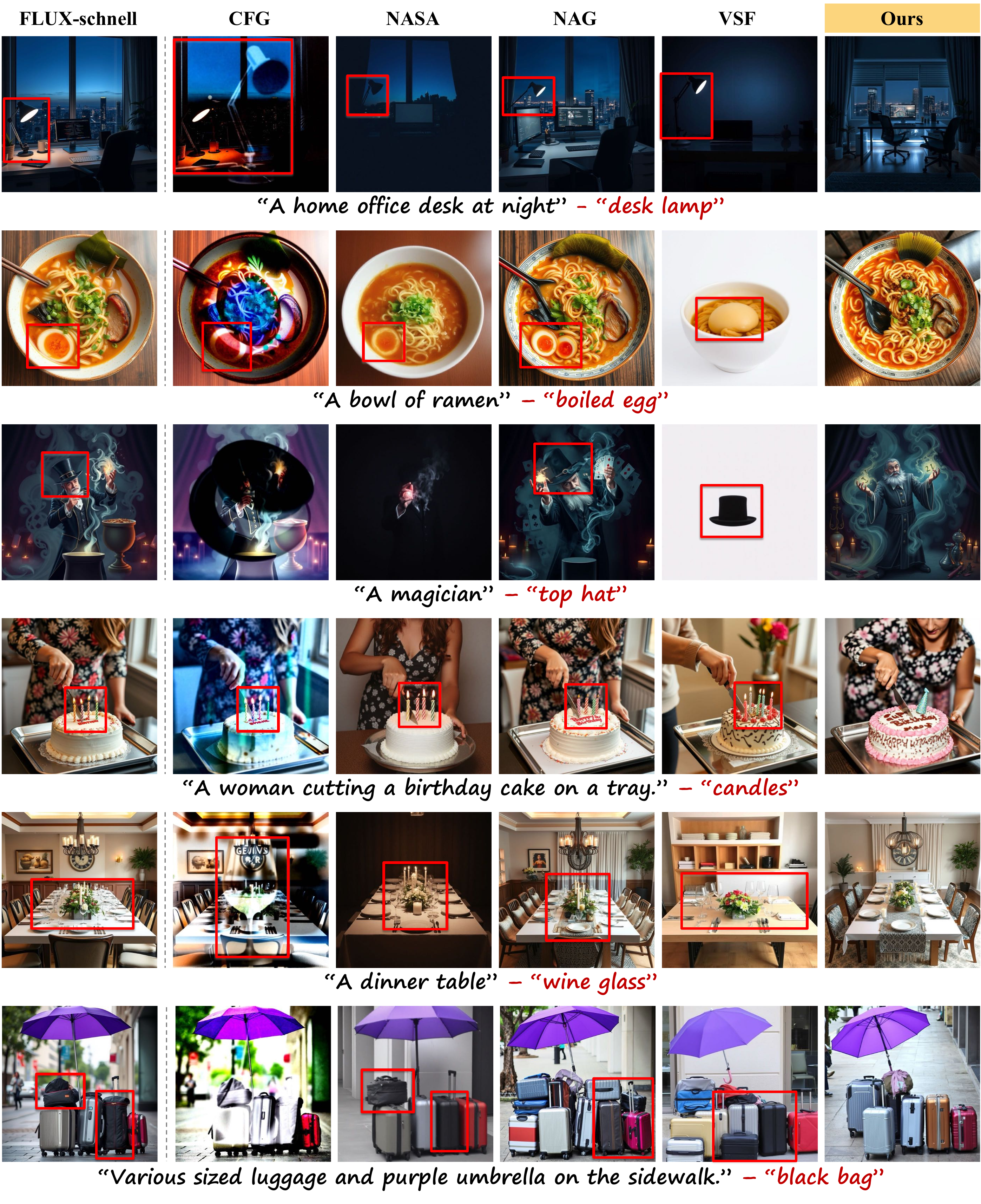}
    \vspace{-2mm}
\end{center}
\caption{\textbf{Additional qualitative comparison with negative guidance methods on FLUX-schnell.}}
\vspace{-1mm}
\label{fig:appendix_schnell_guidance}
\end{figure*}

\begin{figure*}[t]
\begin{center}
   \includegraphics[width=0.85\linewidth]{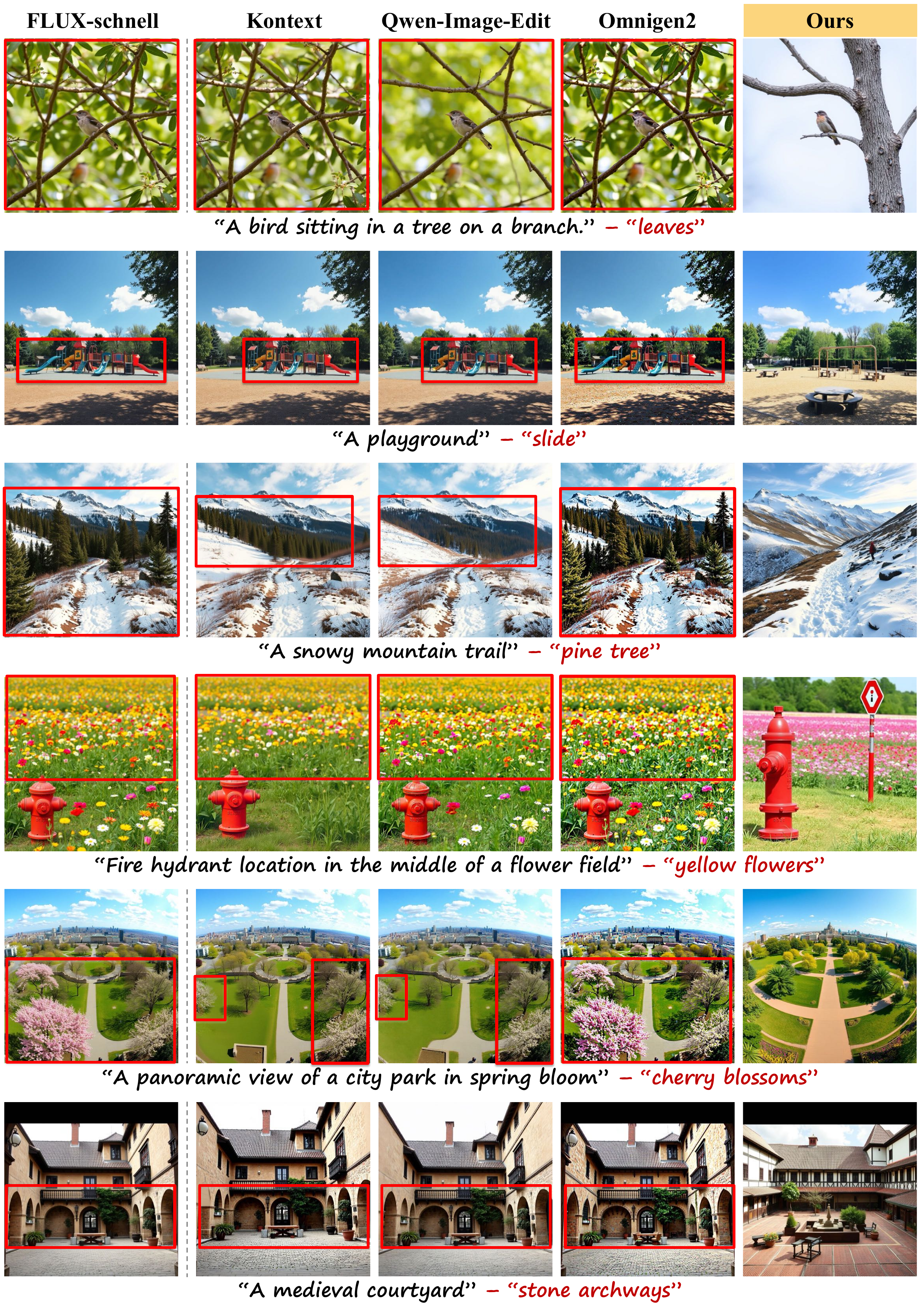}
    \vspace{-2mm}
\end{center}
\caption{\textbf{Additional qualitative comparison with generate-and-edit baselines on FLUX-schnell.}}
\vspace{-1mm}
\label{fig:appendix_schnell_editing}
\end{figure*}

\begin{figure*}[t]
\begin{center}
   \includegraphics[width=0.7\linewidth]{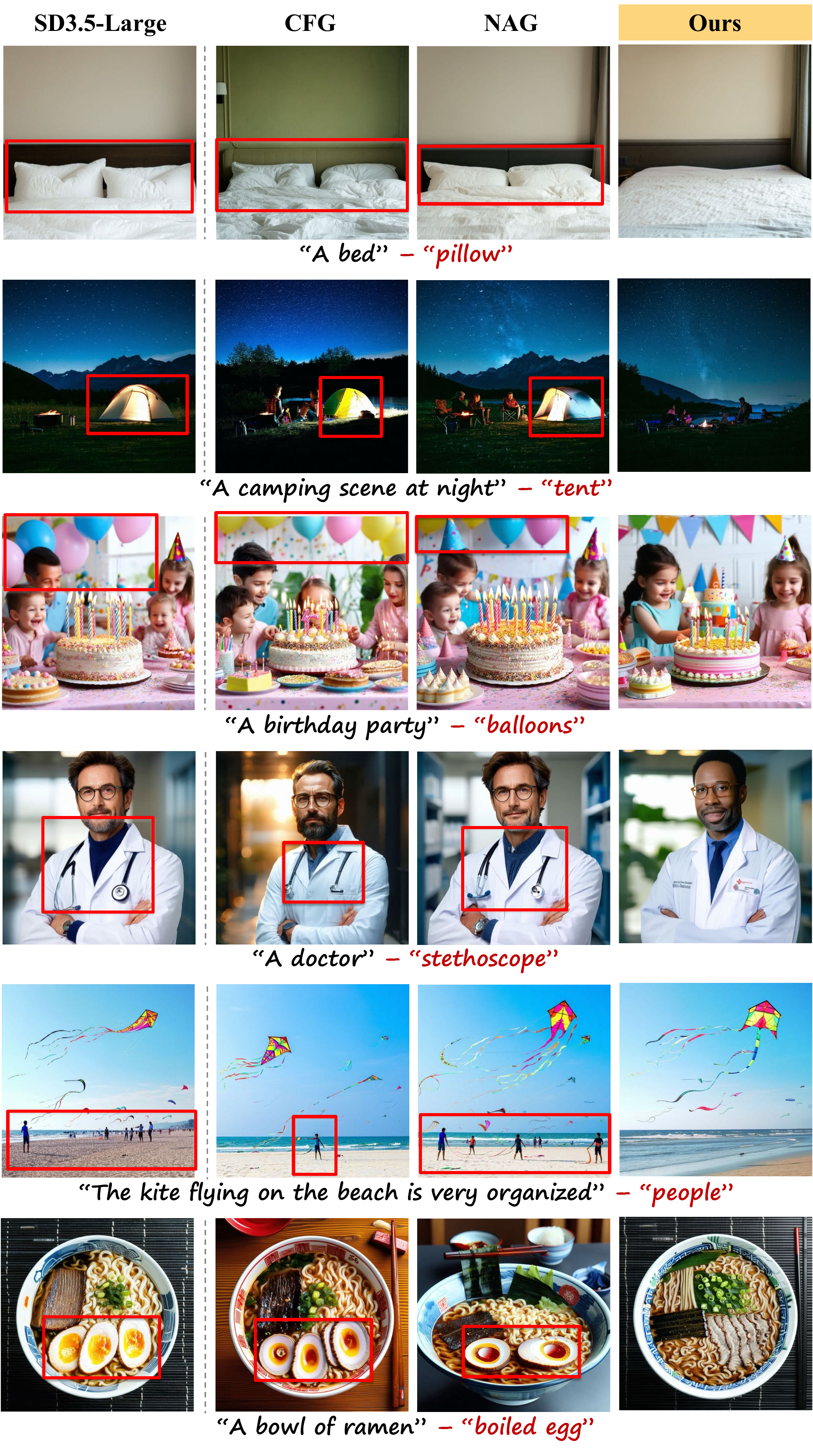}
    \vspace{-2mm}
\end{center}
\caption{\textbf{Additional qualitative comparison with negative guidance methods on SD3.5-Large.}}
\vspace{-1mm}
\label{fig:appendix_sd35_guidance}
\end{figure*}

\begin{figure*}[t]
\begin{center}
   \includegraphics[width=0.85\linewidth]{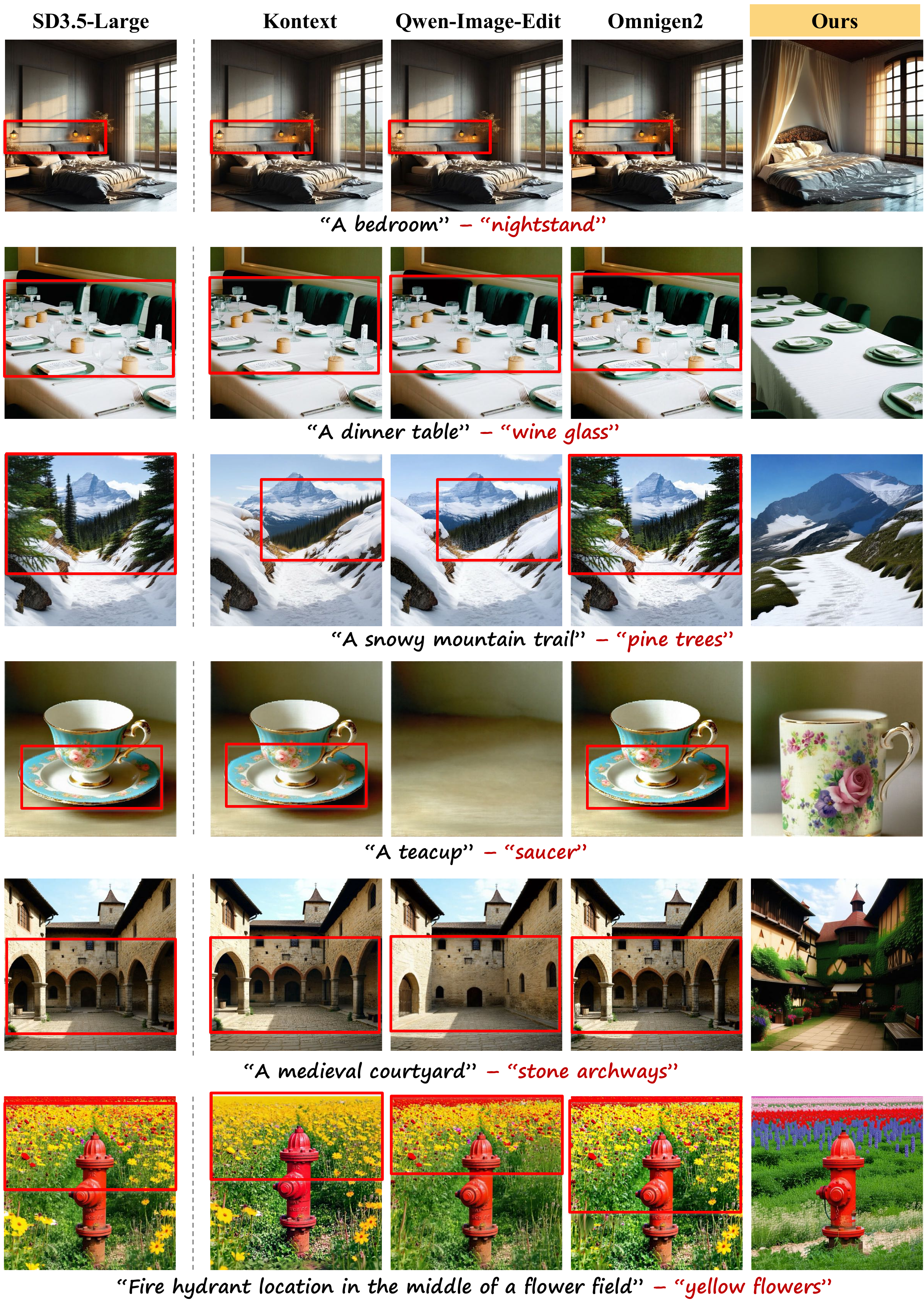}
    \vspace{-2mm}
\end{center}
\caption{\textbf{Additional qualitative comparison with generate-and-edit baselines on SD3.5-Large.}}
\vspace{-1mm}
\label{fig:appendix_sd35_editing}
\end{figure*}

\begin{figure}[t]
\begin{center}
   \includegraphics[width=\linewidth]{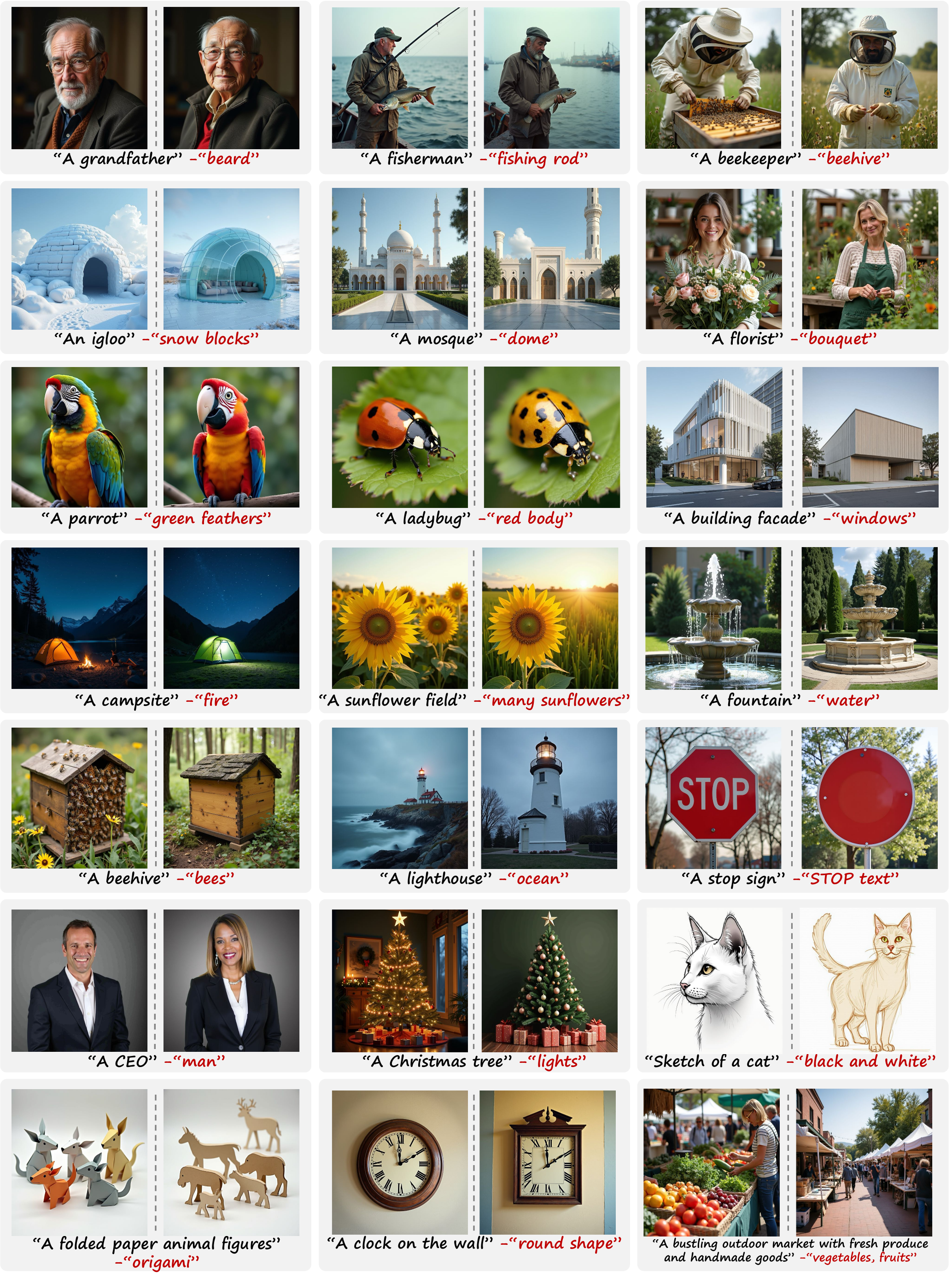}   
    \vspace{-2mm}
\end{center}
\caption{\textbf{Additional qualitative results on suppression scenarios not included in DCS-Bench.} Images without negative guidance are shown on the left of each pair, and images generated with our Orthogonal Negative Guidance are shown on the right. All results are generated using FLUX-dev.
}
\vspace{-1mm}
\label{fig:appendix_general} 
\end{figure}

Figures~\ref{fig:appendix_dev_guidance}--\ref{fig:appendix_sd35_editing} show additional qualitative results on FLUX-dev, FLUX-schnell, and SD3.5-Large. 
Our method successfully suppresses the target concepts across all three models, demonstrating generalization across MM-DiT-based architectures.  
Additional suppression scenarios that are not included in DCS-Bench are shown in Figure~\ref{fig:appendix_general}.

\subsection{Quantitative Results}
\label{sec:appendix_quantitative_results}
\begin{figure}[]
\begin{center}
   \includegraphics[width=\linewidth]{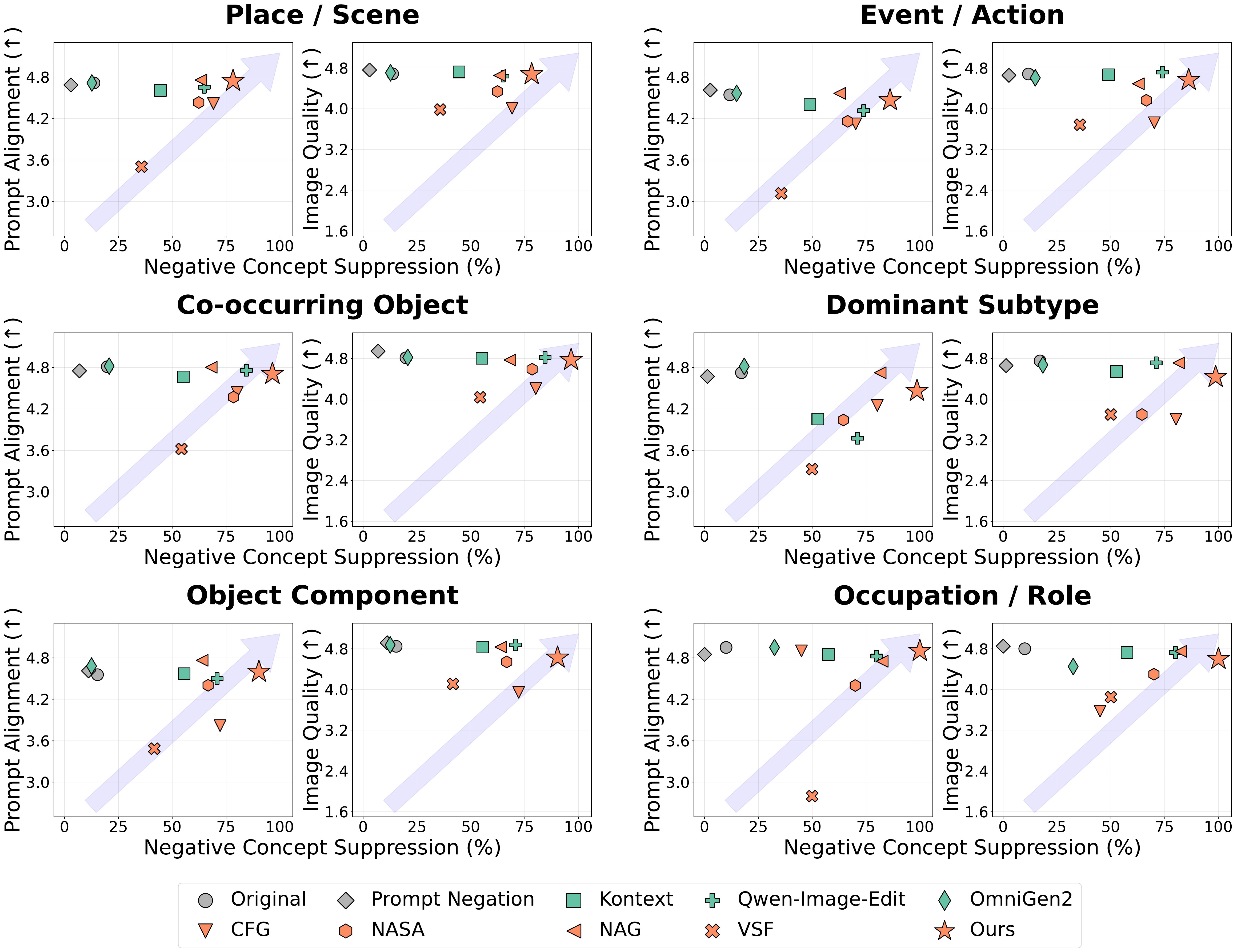}
    \vspace{-2mm}
\end{center}
\caption{\textbf{Per-category quantitative comparison on FLUX-dev.} The six subplot labels correspond to the benchmark categories: Place/Scene = Associated Concept for Place/Scene, Event/Action = Associated Concept for Event/Action, Co-occurring Object = Co-occurring Object for an Object, Dominant Subtype = Dominant Subtype of a Supercategory, Object Component = Associated Component of an Object, and Occupation/Role = Associated Concept for Occupation/Role.}
\vspace{-1mm}
\label{fig:appendix_pareto_category_dev}
\end{figure}

\begin{figure}[]
\begin{center}
   \includegraphics[width=\linewidth]{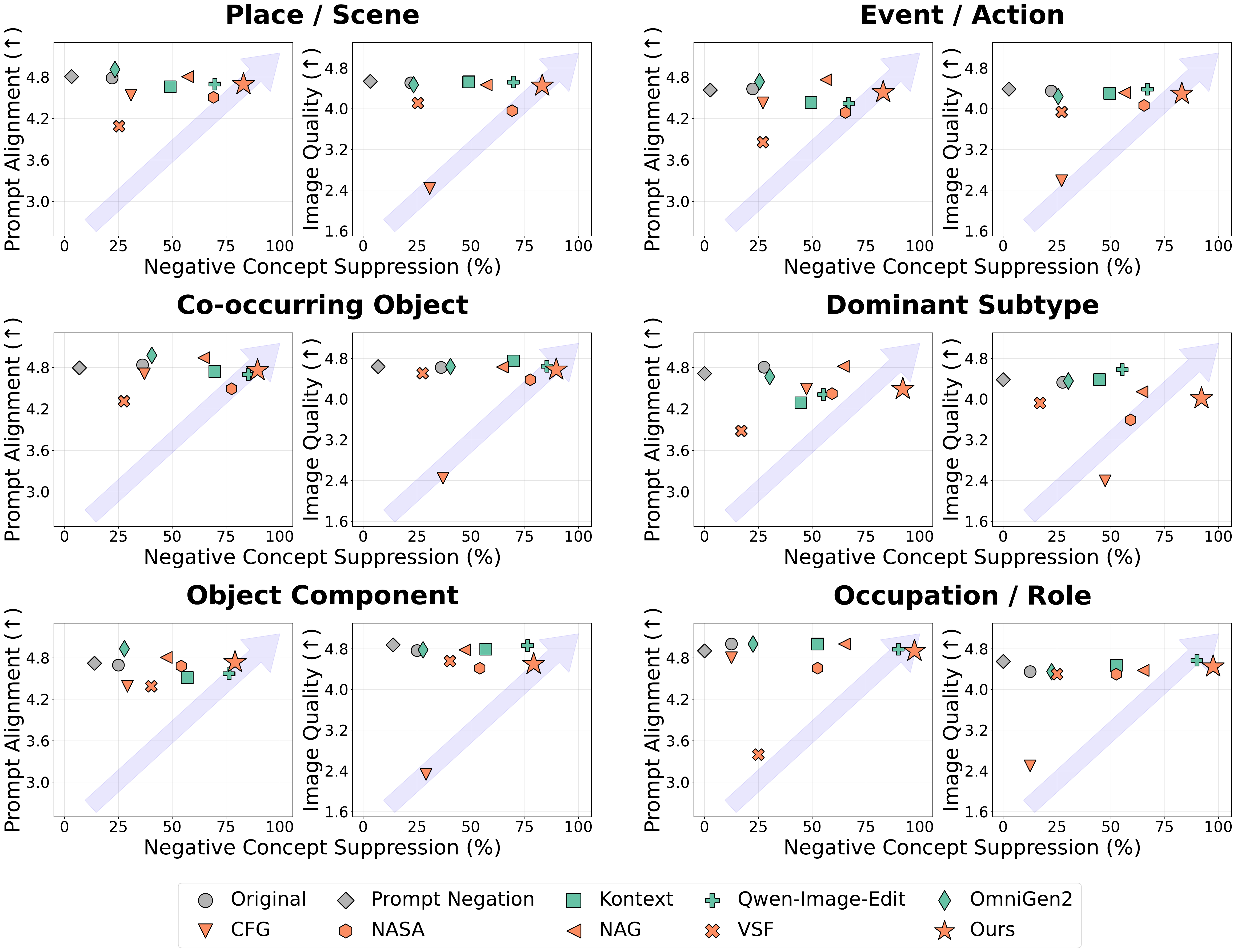}
    \vspace{-2mm}
\end{center}
\caption{\textbf{Per-category quantitative comparison on FLUX-schnell.} The six subplot labels correspond to the benchmark categories: Place/Scene = Associated Concept for Place/Scene, Event/Action = Associated Concept for Event/Action, Co-occurring Object = Co-occurring Object for an Object, Dominant Subtype = Dominant Subtype of a Supercategory, Object Component = Associated Component of an Object, and Occupation/Role = Associated Concept for Occupation/Role.}
\vspace{-1mm}
\label{fig:appendix_pareto_category_schnell}
\end{figure}

Figures~\ref{fig:appendix_pareto_category_dev} and~\ref{fig:appendix_pareto_category_schnell} present per-category results on FLUX-dev and FLUX-schnell, respectively.
Our method achieves the highest Negative Concept Suppression across all categories while maintaining competitive Prompt Alignment and Image Quality.

Tables~\ref{tab:flux_results} and~\ref{tab:schnell_results} further break down results by data source (LLM-generated vs.\ COCO-derived).
The performance of our method remains consistent across both subsets, and aligns with the combined results reported in the main paper.


  \begin{table}[]
  \centering
  \caption{\textbf{Quantitative comparison on FLUX-dev.} Suppression: Negative Concept Suppression~(\%), Align: Prompt Alignment, Quality: Image Quality.
  Higher is better for all metrics.
  \textbf{Bold} and \underline{Underlined} indicate the best and
  second-best results, respectively.}
  \label{tab:flux_results}
  \resizebox{\textwidth}{!}{%
  \begin{tabular}{l|ccc|ccc|ccc}
  \toprule
  \multirow{2}{*}{Method} & \multicolumn{3}{c|}{LLM-generated} & \multicolumn{3}{c|}{COCO-derived} &
  \multicolumn{3}{c}{Combined} \\
  & Suppression(\%)  & Align  & Quality  & Suppression(\%)  & Align  &
  Quality  & Suppression(\%)  & Align  & Quality  \\
  \midrule
  FLUX-dev   & 13.25 & 4.83 & \underline{4.83} & 15.50 & 4.54 & \underline{4.63} & 14.38 & 4.69 & \underline{4.73} \\
  Prompt Negation    & 3.50 & \textbf{4.86} & \textbf{4.84} & 4.25 & 4.49 & \textbf{4.70} & 3.88 & 4.68 & \textbf{4.77} \\
  \midrule
  Kontext    & 49.50 & 4.75 & 4.82 & 49.50 & 4.29 & 4.61 & 49.50 & 4.52 & 4.71 \\
  Qwen-Image-Edit  & \underline{71.00} & 4.74 & 4.81 & 72.50 & 4.26 & 4.62 & \underline{71.75} & 4.50 & 4.72 \\
  OmniGen2   & 15.50 & \underline{4.84} & 4.76 & 16.25 & \underline{4.59} & \underline{4.63} & 15.88 & \textbf{4.71} & 4.70 \\
  \midrule
  CFG        & 68.75 & 4.43 & 4.11 & \underline{73.50} & 4.18 & 3.71 & 71.12 & 4.30 & 3.91 \\
  NASA       & 64.25 & 4.49 & 4.47 & 69.00 & 4.14 & 4.11 & 66.62 & 4.32 & 4.29 \\
  NAG        & 60.75 & 4.81 & 4.71 & 72.50 & \textbf{4.62} & 4.60 & 66.62 & \underline{4.71} & 4.66 \\
  VSF        & 42.25 & 3.04 & 3.77 & 39.75 & 3.72 & 4.03 & 41.00 & 3.38 & 3.90 \\
  Ours       & \textbf{85.25} & 4.79 & 4.75 & \textbf{88.50} & 4.49 & 4.51 & \textbf{86.88} & 4.64 & 4.63 \\
  \bottomrule
  \end{tabular}
  }
  \end{table}

\begin{table}[]
  \centering
  \caption{\textbf{Quantitative comparison on FLUX-schnell.} Suppression: Negative Concept Suppression~(\%), Align: Prompt Alignment, Quality: Image Quality.
    Higher is better for all metrics.
    \textbf{Bold} and \underline{Underlined} indicate the best and
    second-best results, respectively.}
  \label{tab:schnell_results}
  \resizebox{\textwidth}{!}{%
  \begin{tabular}{l|ccc|ccc|ccc}
  \toprule
  \multirow{2}{*}{Method} & \multicolumn{3}{c|}{LLM-generated} & \multicolumn{3}{c|}{COCO-derived} &
  \multicolumn{3}{c}{Combined} \\
  & Suppression(\%)  & Align  & Quality  & Suppression(\%)  & Align  &
  Quality  & Suppression(\%)  & Align  & Quality  \\
  \midrule
  FLUX-schnell   & 16.75 & 4.90 & \textbf{4.66} & 32.25 & 4.62 & 4.31 & 24.50 & 4.76 & 4.48 \\
  Prompt Negation    & 3.25 & 4.87 & 4.65 & 5.00 & 4.60 & 4.38 & 4.12 & 4.74 & \underline{4.51} \\
  \midrule
  Kontext    & 46.75 & 4.83 & 4.63 & 58.50 & 4.35 & \underline{4.40} & 52.62 & 4.59 & \underline{4.51} \\
  Qwen-Image-Edit  & \underline{71.00} & 4.81 & \textbf{4.66} & \underline{72.25} & 4.40 & \textbf{4.43} &
  \underline{71.62} & 4.61 & \textbf{4.55} \\
  OmniGen2   & 19.25 & \underline{4.90} & 4.55 & 35.50 & \underline{4.78} & 4.35 & 27.38 & \underline{4.84} & 4.45 \\
  \midrule
  CFG        & 25.50 & 4.65 & 2.48 & 37.25 & 4.41 & 2.44 & 31.38 & 4.53 & 2.46 \\
  NASA       & 62.75 & 4.59 & 4.20 & 70.00 & 4.34 & 3.94 & 66.38 & 4.47 & 4.07 \\
  NAG        & 52.25 & \textbf{4.95} & 4.62 & 64.25 & \textbf{4.70} & 4.28 & 58.25 & \textbf{4.83} & 4.45 \\
  VSF        & 21.50 & 3.87 & 4.13 & 31.75 & 4.21 & 4.18 & 26.62 & 4.04 & 4.16 \\
  Ours       & \textbf{84.25} & 4.83 & 4.53 & \textbf{86.25} & 4.51 & 4.26 & \textbf{85.25} & 4.67 & 4.39 \\
  \bottomrule
  \end{tabular}
  }
  \end{table}

\clearpage
\subsection{Comparison with Prompt-Level Negation}
\label{sec:appendix_prompt_negation}

\begin{table}[h]
\centering
\small
\caption{\textbf{Negative Concept Suppression results on FLUX-dev.}}

\label{tab:prompt_negation_comparison}
\begin{tabular}{lc}
\toprule
Method & Negative Concept Suppression ($\uparrow$) \\
\midrule
FLUX-dev & 14.38\\
Prompt Negation & 3.88 \\
Ours &  86.88 \\
\bottomrule
\end{tabular}
\end{table}

\begin{figure}[h]
\begin{center}
   \includegraphics[width=0.55\linewidth]{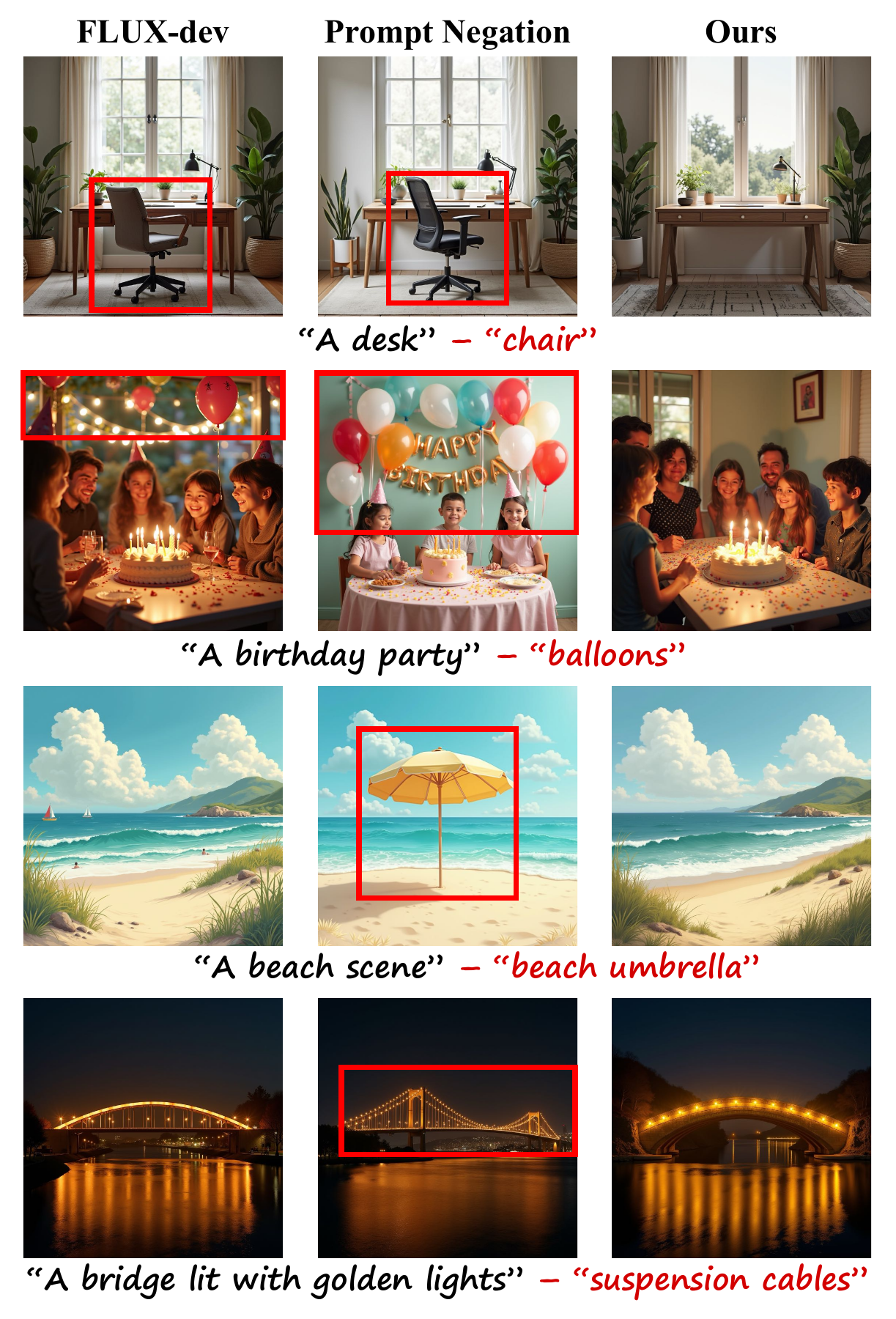}
\end{center}
\caption{\textbf{Qualitative Comparison with prompt-level negation on FLUX-dev.}}
\label{fig:appendix_prompt_negation}
\end{figure}

Prompt-level negation is often unreliable for suppressing unwanted concepts and may even lead to the generation of the suppression target~\cite{li2024get,alhamoud2025vision}. 
To examine this behavior, we evaluate a simple prompt-level negation baseline, denoted as \textit{Prompt Negation}, which appends the phrase ``without \{suppression target\}'' to the prompt. 
Quantitative results are reported in Table~\ref{tab:prompt_negation_comparison}.

Figure~\ref{fig:appendix_prompt_negation} shows qualitative comparisons.
In some cases (rows 1 and 2), prompt-level negation fails to remove the suppression target despite the negation phrase. 
In other cases (rows 3 and 4), it even generates the suppression target, although the image generated from the original prompt does not contain it. 
In contrast, our method successfully suppresses the target concept while preserving prompt alignment.

\subsection{Ablation on Attention Features for Negative Guidance}
\label{sec:appendix_ablation_attention_component}

\begin{figure}[h]
    \centering
    \includegraphics[width=\linewidth]{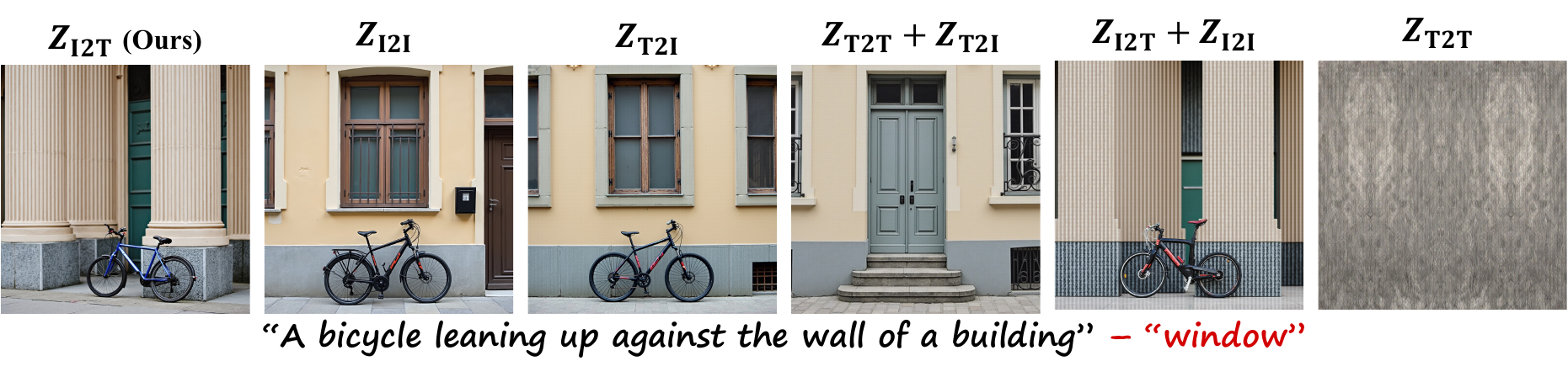}
    \caption{\textbf{Ablation on attention features for negative guidance.}
     Each column applies orthogonal guidance to a different attention output component.
    From left to right: $Z_{\mathrm{I2T}}$ (ours), $Z_{\mathrm{I2I}}$, $Z_{\mathrm{I2T}} + Z_{\mathrm{I2I}}$, $Z_{\mathrm{T2T}}$, $Z_{\mathrm{T2I}}$, $Z_{\mathrm{T2T}} + Z_{\mathrm{T2I}}$.
    Only $Z_{\mathrm{I2T}}$ removes suppression target `window' while preserving both prompt alignment and image quality.}
    \label{fig:ablation_attention_component}
\end{figure}

As shown in Eq.(\ref{eqn:Orthogonal Guidance}), our method applies orthogonal negative guidance to the image-to-text attention output $Z_{\mathrm{I2T}}$ in MM-DiT attention layers. This design is motivated by prior work showing that image-to-text attention captures semantic alignment between the image and text modalities~\cite{kim2025reflex, shin2025exploring}. To empirically evaluate the choice of attention feature for applying negative guidance, we conduct an ablation study by applying orthogonal negative guidance to different attention components: $Z_{\mathrm{I2T}}$, $Z_{\mathrm{I2I}}$, $Z_{\mathrm{I2T}} + Z_{\mathrm{I2I}}$, $Z_{\mathrm{T2T}}$, $Z_{\mathrm{T2I}}$, and $Z_{\mathrm{T2T}} + Z_{\mathrm{T2I}}$. A qualitative result is shown in Figure~\ref{fig:ablation_attention_component}. In this example, applying negative guidance to $Z_{\mathrm{I2I}}$, $Z_{\mathrm{T2I}}$, or $Z_{\mathrm{T2T}} + Z_{\mathrm{T2I}}$ fails to suppress the target concept `window’. Applying negative guidance to $Z_{\mathrm{I2T}} + Z_{\mathrm{I2I}}$ or $Z_{\mathrm{T2T}}$ results in severe image quality degradation. Notably, when applied to $Z_{\mathrm{T2T}} + Z_{\mathrm{T2I}}$, the guidance incorrectly suppresses the object `bicycle’ specified in the prompt instead of the `window’ specified in the negative prompt. In contrast, our method, which applies negative guidance to $Z_{\mathrm{I2T}}$, effectively suppresses the target concept while maintaining strong prompt alignment and image quality. These results indicate that the effectiveness of negative guidance depends on the choice of attention feature, with $Z_{\mathrm{I2T}}$ performing best in our ablation.

\clearpage
\section{Additional Application Results}
\label{sec:appendix_controllable_multi_concept}

\subsection{Additional Results on Adjustable Concept Suppression}
\label{sec:appendix_adjustable}

\begin{figure}[]
\begin{center}
   \includegraphics[width=0.85\linewidth]{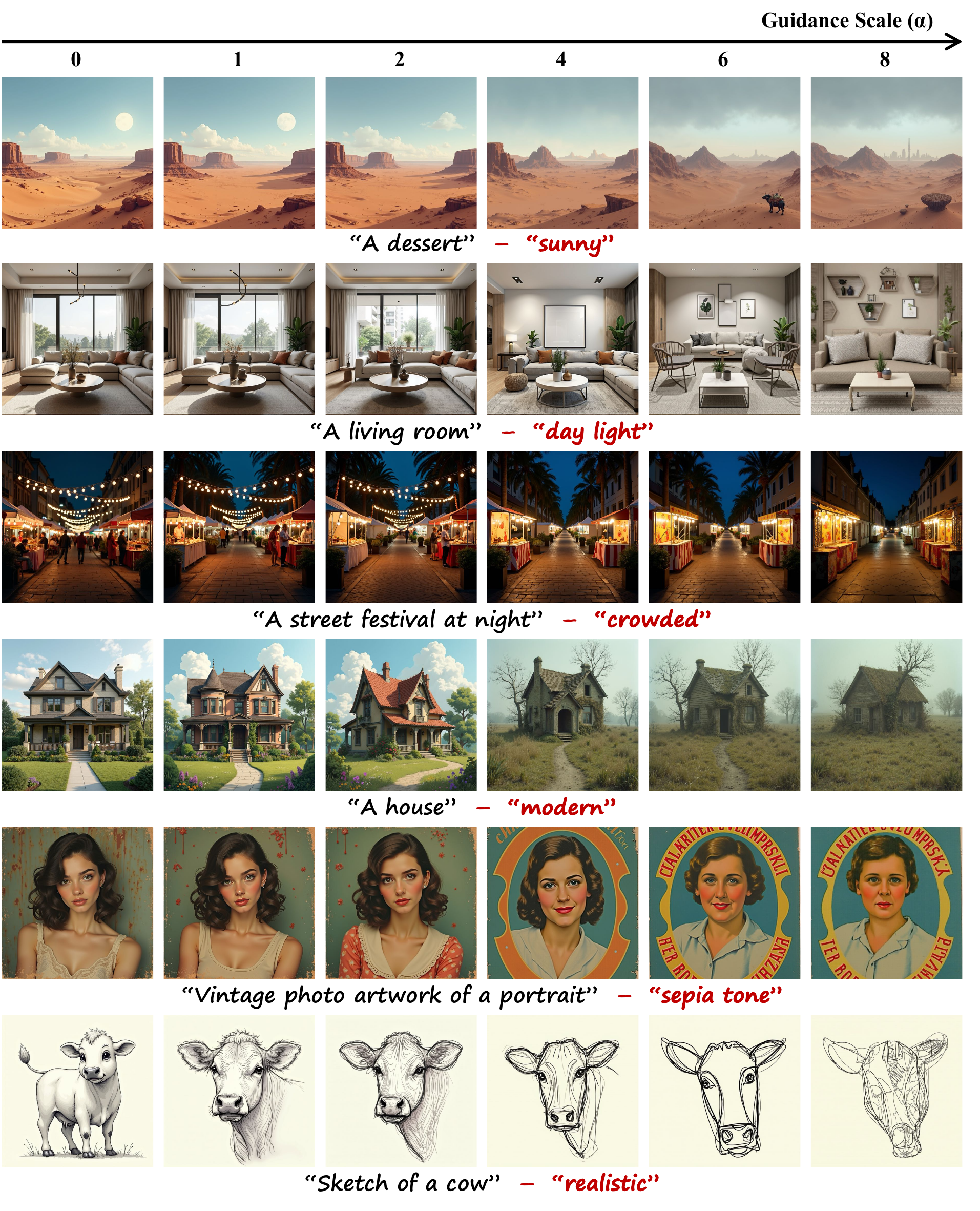}  
    \vspace{-2mm}
\end{center}
\caption{
  \textbf{Additional examples of adjustable concept suppression.}
  $\alpha=0$ corresponds to generation without negative guidance.
}
\vspace{-1mm}
\label{fig:appendix_control_suppression} 
\end{figure}

Figure~\ref{fig:appendix_control_suppression} provides additional examples of adjustable concept suppression.

\clearpage
\subsection{Additional Results on Multi-Concept Suppression}
\label{sec:appendix_multi_concept}
\begin{figure}[]
\begin{center}
   \includegraphics[width=0.8\linewidth]{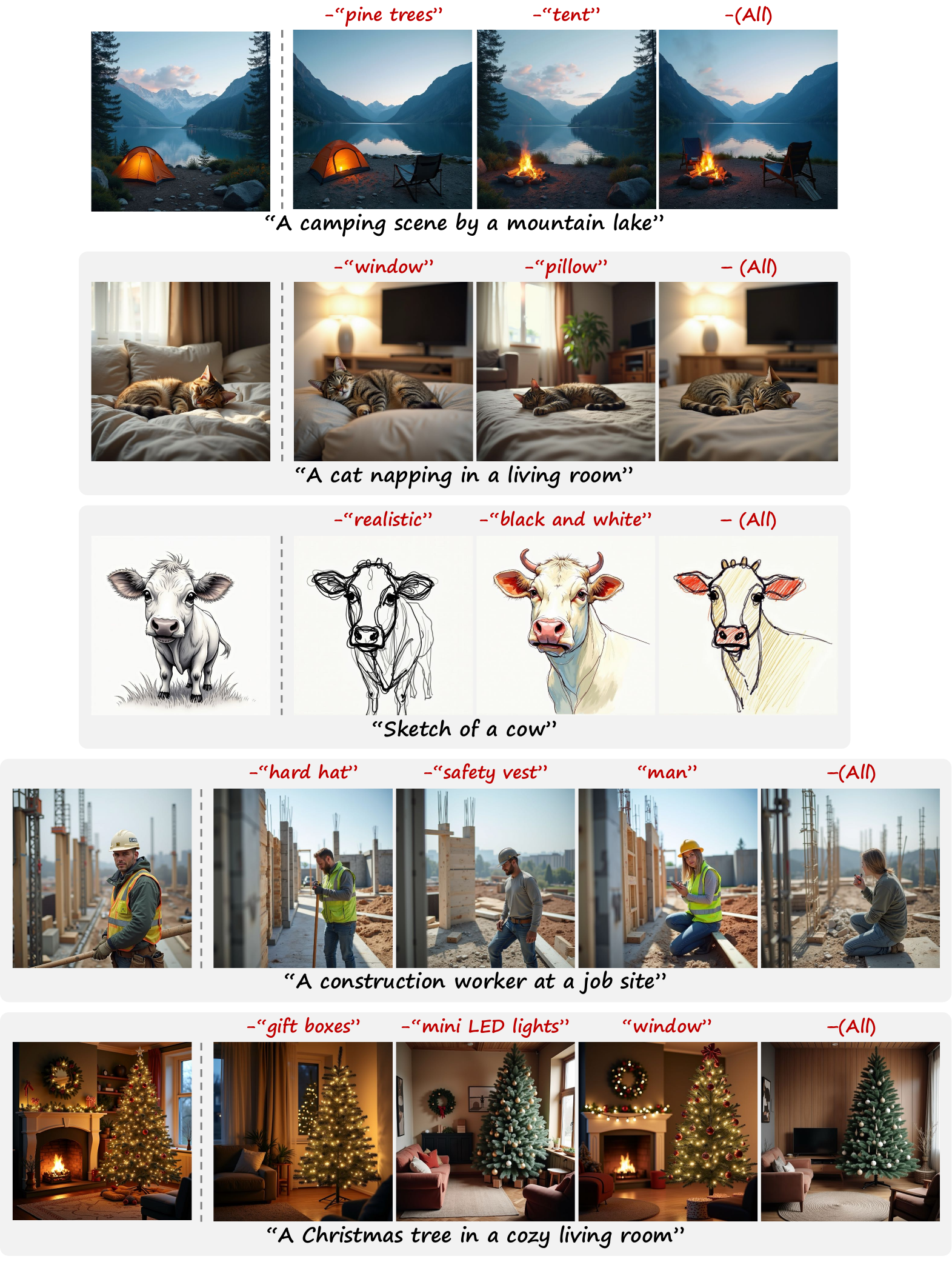}  
    \vspace{-2mm}
\end{center}
\caption{
  \textbf{Additional examples of multi-concept suppression.}
  The leftmost images correspond to generation without negative guidance.
}
\vspace{-1mm}
\label{fig:appendix_multi_concept} 
\end{figure}

Figure~\ref{fig:appendix_multi_concept} provides additional examples of multi-concept suppression.

\clearpage
\section{Full List of Suppression Scenarios in DCS-Bench}
\label{sec:appendix_full_pair_list}
Tables~\ref{tab:pairs_place}--\ref{tab:pairs_hypernym} list all 200 concept suppression scenarios in DCS-Bench, grouped by category.

\begin{table}[h]
\centering
\caption{\textbf{Associated Concept for Place/Scene (77 pairs) -- Part 1 of 2.}}
\label{tab:pairs_place}
\scriptsize
\begin{tabular}{p{0.65\linewidth}p{0.25\linewidth}}
\toprule
Prompt & Suppression Target \\
\midrule
A bathroom & bathtub \\
A living room & sofa \\
A bedroom & nightstand \\
A busy train station & arrival board \\
A beach scene & beach umbrella \\
A laundry room & laundry basket \\
A cruise ship deck & deck chairs \\
A snowy mountain trail & pine trees \\
A playground & slide \\
A hotel lobby & chandelier \\
A black cat on a moonlit rooftop & chimney \\
A bright blue sky over a meadow at sunrise & clouds \\
A calm beach at golden hour & footprints \\
A bustling outdoor caf\'{e} on a European street & cobblestone pavement \\
A bustling urban street during rush hour & traffic lights \\
A canyon overlook at sunset & orange cliffs \\
A classic movie theater interior & red curtains \\
A construction site at dusk & tower cranes \\
A crowded marketplace at midday & canopies \\
A cozy bedroom at twilight & bedside lamp \\
A fish market at dawn & hanging lamps \\
A futuristic server room aisle & blue indicator LEDs \\
A hidden waterfall in a jungle & mist \\
A medieval courtyard & stone archways \\
A mountain village lit by lanterns & stone steps \\
A narrow European alley at midday & window shutters \\
A quaint wooden cottage in a foggy forest clearing & moss \\
A quiet meditation garden & stepping stones \\
A riverside promenade at sunset & lampposts \\
A small-town diner & neon sign \\
A snow-covered village at night & wooden fences \\
A sports car on a racetrack & safety barriers \\
A starry night above a quiet desert & Milky Way \\
A swimming pool at midday & deck chairs \\
A vintage record store & posters \\
An amusement park at night & ferris wheel \\
A construction area with a road sign for the near street. & cone \\
A black and white photo of a street with many cars parked on the side. & buildings \\
A man is walking a dog on his bike through a crosswalk. & cars \\
\bottomrule
\end{tabular}
\end{table}

\begin{table}[h]
\centering
\caption{\textbf{Associated Concept for Place/Scene (77 pairs) -- Part 2 of 2.}}
\scriptsize
\begin{tabular}{p{0.65\linewidth}p{0.25\linewidth}}
\toprule
Prompt & Suppression Target \\
\midrule
A road filled with snow and traffic lights. & trees \\
Rows of cars parked on the side of a city street going uphill. & trees \\
A very tall clock tower sitting in the middle of a city. & cars \\
A locomotive crossing on a street with the arm down to stop traffic as the train is passing through. & car \\
The inside of a building with a sculpture of baseball bats and baseballs hanging from the ceiling. & windows \\
A living room with television and lights turned on. & sofa \\
A minimalist yoga studio & yoga mats \\
A view of a sign that says Western Av, but its sideways and on the ground. & grass \\
A train on the tracks out in the country & grass \\
A couple of horses are on the side of the road. & power lines \\
A road with a traffic light next to a bush and a house. & power lines \\
A Cave sign is next to a restaurant and lounge. & table \\
Company breakroom with black and white tiled floor and a vending machine. & chair \\
A modern bathroom is designed to be useful. & plant \\
A hotel room filled with beige and blue furniture. & table \\
The scene is three large racks of skis on a mountain. & clouds \\
A body of water, with a background of mountains. & clouds \\
A very cluttered room has lots of suitcases including a red trunk on the floor. & chair \\
A surfboard is being displayed in an office. & table \\
A herd of cattle grazing on top of a grass covered beach. & waves \\
A tree lined river in front of a hill. & stone \\
A white bench on a cliff by the sea shore on a sunny day. & grass \\
A boat sitting on a beach next to an ocean. & hills \\
A bookstore reading nook & stacked books \\
A cozy living room with a cat napping & cushions \\
A flower shop storefront & vases \\
A glass greenhouse full of exotic plants & pots \\
A glassblowing studio & metal tongs \\
A high-tech laboratory interior & beakers \\
A mechanic's garage at night & tool chest \\
A professional photography studio & softbox lights \\
A rustic wine cellar & wooden barrels \\
A small art gallery & wooden bench \\
A room with bookshelves, a wood floor, and bean bag chairs. & window \\
A street sign hanging from the ceiling in a room with a fan on the wall. & door \\
A wooden bench with an iron frame sitting on a boardwalk. & lamppost \\
An earth tone kitchen has a gas range, tan countertops and oak cabinets. & sink \\
a red bag is on the floor in a room & curtain \\
\bottomrule
\end{tabular}
\end{table}

\begin{table}[h]
\centering
\caption{\textbf{Associated Concept for Event/Action (47 pairs).}}
\label{tab:pairs_event}
\scriptsize
\begin{tabular}{p{0.65\linewidth}p{0.25\linewidth}}
\toprule
Prompt & Suppression Target \\
\midrule
A birthday party & balloons \\
A wedding ceremony & veil \\
A woman packing for a trip & suitcase \\
A camping scene at night & tent \\
A holiday dinner & table runner \\
A camper cooking outdoors & portable stove \\
A cooking class & cutting board \\
A couple having a picnic & picnic blanket \\
A student during online class & notepad \\
A waiter setting a table & cloth napkin \\
A ceramics workshop in progress & aprons \\
A person walking in snow & footprints \\
A protest scene & placards \\
A couple having dinner & candlelight \\
A desert caravan & camels \\
A child blowing soap bubbles & grass \\
A jazz band on stage & amplifiers \\
A tram stop on a rainy morning & puddles \\
An open-air barbecue grill & smoke \\
there giraffes in a penned area standing outside & tree \\
A herd of sheep grazing on a lush green field. & sky \\
A woman feeding a giraffe with her hand. & fence \\
Bunch of swans in the water boats on the other side. & buildings \\
a bird standing close to a parked car & fallen leaves \\
A view of a wet day from indoors. & buildings \\
Several pictures of people using tablets and laptops. & plant \\
The kite flying on the beach is very organized & people \\
There are lot of people rowing on the boat. & mountains \\
a couple of helicopters are in the sky & mountains \\
Someone is protesting the financial control of Wall Street. & signs \\
A big full view of several people gathering. & stage \\
A cat that is checking into its Twitter account to post a tweet. & laptop \\
A guitarist performing on stage & microphone \\
A scientist giving a talk & projector screen \\
A group of hikers ascending a rocky mountain trail & backpacks \\
A pottery display at a craft fair & ceramic plates \\
A small rowboat tied to a dock & rope \\
A multitude of people cross the street in front of streetcar. & crosswalk \\
A large rounded mirror to help people see past certain angles. & chair \\
A woman holds her skis as her friends go down the trail. & backpack \\
Several people in boats full of vegetables and fruit, with a market in the background. & hats \\
A woman pulling her suitcase on the sidewalk along a brick paved street & coat \\
A man swinging a racquet on a tennis court. & ball \\
There is a paper with a drawing an a nokia phone on top of it & pencil \\
a person riding a motorcycle near a stop sign & backpack \\
Playing baseball in a local field with Dad & bat \\
A man a woman and a dog walk in the snow. & leash \\
\bottomrule
\end{tabular}
\end{table}

\begin{table}[h]
\centering
\caption{\textbf{Co-occurring Object for an Object (29 pairs).}}
\label{tab:pairs_independent}
\scriptsize
\begin{tabular}{p{0.65\linewidth}p{0.25\linewidth}}
\toprule
Prompt & Suppression Target \\
\midrule
A desk & chair \\
A piano & sheet music \\
A dinner table & wine glass \\
A Christmas tree & gift boxes \\
A kitchen island & fruit bowl \\
A fireplace mantel & photo frame \\
A food truck & condiment shelf \\
A patio table & centerpiece pot \\
A set of park benches near a lamp post & path \\
A bed set up on the wooden floor of a large wooden house & nightstand \\
A modern office workspace with multiple monitors & keyboard \\
a keyboard is laying on a wood desk & monitor \\
A laptop computer that is on a desk. & smartphone \\
A cheese board & grapes \\
A kitchen drying rack & dish towel \\
A taco tray & lime wedge \\
A teacup & saucer \\
A home office desk at night & desk lamp \\
A colorful bath curtain in blue stripes and a bath and hand towel to match & stool \\
A black cat sitting on top of a bathroom sink. & mirror \\
A white automatic dishwasher sitting under a microwave oven. & plates \\
A television next to a couple of desktop computers on a desk. & speaker \\
some onions in a silver bowl and some meat in another and a spoon & parsley \\
A table setting of colorful food. & wine \\
A necklace that is under a hair clip. & earrings \\
a polar bear partly submerged in water & ice \\
A grouo of four snowboards out in front of a red colors building. & snow \\
A vintage wood stove used for cooking or baking. & baskets \\
A picture of a blender sitting on a counter. & bowl \\
\bottomrule
\end{tabular}
\end{table}

\begin{table}[h]
\centering
\caption{\textbf{Dominant Subtype of a Supercategory (19 pairs)}.}
\label{tab:pairs_hypernym}
\scriptsize
\begin{tabular}{p{0.65\linewidth}p{0.25\linewidth}}
\toprule
Prompt & Suppression Target \\
\midrule
A vase of flowers is partially in the dark. & roses \\
a bunch of animals that are in a field & deer \\
Two giraffes and several other large animals roam a tropical zoo paddock. & elephants \\
some large wooden forks and other items on display & spoons \\
A variety of fruit placed in a wicker basket & apple \\
Two small animal figurines on the keypad of a cellphone. & orange animal figurine \\
Fire hydrant location in the middle of a flower field & yellow flowers \\
A picture of stuffed animals that are dressed up and in a box. & bear \\
A panoramic view of a city park in spring bloom & cherry blossoms \\
This is a picture of a white refrigerator with several magnets on the doors. & hearts \\
A plate with an enormous portion of gravy topped food and a side of veggies. & broccoli \\
A plate of food with a hot dog and a carton of milk. & french fries \\
there are two plates of different foods on the table & potato \\
A plate of food is arranged with fruit and vegetables. & tomato \\
A collection of photos of Japanese dishes and soup. & egg \\
A plate with a sandwich and fries on it in front of a glass of beer. & burger \\
Various sized luggage and purple umbrella on the sidewalk. & black bag \\
The back of the vehicle is loaded with ski equipment. & helmet \\
Here is a trunk with stationary and other memorabilia tucked inside. & book \\
\bottomrule
\end{tabular}
\end{table}

\begin{table}[h]
\centering
\caption{\textbf{Associated Component of an Object (18 pairs).}}
\label{tab:pairs_component}
\scriptsize
\begin{tabular}{p{0.65\linewidth}p{0.25\linewidth}}
\toprule
Prompt & Suppression Target \\
\midrule
A bowl of ramen & boiled egg \\
A bird sitting in a tree on a branch. & leaves \\
A bulletin board & pushpins \\
A bed & pillow \\
A cat curled up on a bed for a nap. & pillow \\
A painting of a small child with a skateboard on a slope & flower \\
a desktop setup with a laptop and two monitors & keyboard \\
A fruit basket & orange \\
A salad plate & cherry tomatoes \\
A bridge lit with golden lights & suspension cables \\
a close up of a sandwich on a plate & lettuce \\
A supreme pizza with two slices taken out. & pepperoni \\
There is a pizza with many toppings on it & basil \\
There are many different types of donuts in the box. & chocolate \\
The tray has two large hot dogs and one regular hot dog. & cheese \\
A woman cutting a birthday cake on a tray. & candles \\
A Christmas wreath covered in red bows fastened to a door. & red balls \\
A bicycle leaning up against the wall of a building & window \\
\bottomrule
\end{tabular}
\end{table}

\begin{table}[h]
\centering
\caption{\textbf{Associated Concept for Occupation/Role (10 pairs).}}
\label{tab:pairs_occupation}
\scriptsize
\begin{tabular}{p{0.65\linewidth}p{0.25\linewidth}}
\toprule
Prompt & Suppression Target \\
\midrule
A doctor & stethoscope \\
A magician & top hat \\
A businesswoman walking & handbag \\
A painter outdoors & easel \\
A DJ performing at a club & headphones \\
A jogger in the city & smartwatch \\
A game streamer & LED lights \\
A street vendor & umbrella \\
Group of businessman discussing something important & glasses \\
Three workers stand next to each other with their baked goods behind them. & aprons \\
\bottomrule
\end{tabular}
\end{table}

\end{document}